\documentclass{article}

\usepackage[preprint]{corl_2026} % Uncomment for pre-prints (e.g., arxiv); This is like ``final'', but will remove the CORL footnote.
\usepackage[utf8]{inputenc} % allow utf-8 input
\usepackage[T1]{fontenc}    % use 8-bit T1 fonts
\usepackage{hyperref}       % hyperlinks
\usepackage{url}            % simple URL typesetting
\usepackage{booktabs}       % professional-quality tables
\usepackage{amsfonts}       % blackboard math symbols
\usepackage{nicefrac}       % compact symbols for 1/2, etc.
\usepackage{microtype}      % microtypography
\usepackage{xcolor}         % colors
\usepackage{times}

\usepackage{multirow}    
\usepackage{makecell}        
\usepackage{graphicx}    
\usepackage{amsmath} 
\usepackage{booktabs}
\usepackage{tabularx} 
\usepackage{amssymb} 
\usepackage{caption}
\usepackage[table]{xcolor}
\usepackage{xcolor}
\usepackage{svg}
\usepackage{wrapfig}
\definecolor{deepPink}{RGB}{255, 20, 147}

\title{BiDexGrasp: Coordinated Bimanual Dexterous Grasps across Object Geometries and Sizes}

\author{
    \small 
    \textbf{Mu Lin}$^{1,3,*}$,
    \textbf{Yi-Lin Wei}$^{1,*}$,
    \textbf{Jiaxuan Chen}\textsuperscript{1}, 
    \textbf{Yuhao Lin}\textsuperscript{1},
    \textbf{Shuoyu Chen}\textsuperscript{1},
    \textbf{Zhizhao Liang}\textsuperscript{1},
    \textbf{Jiangran Lyu}\textsuperscript{2}, \\ 
    \small 
    \textbf{Jiayi Chen}\textsuperscript{2},
    \textbf{Xiaoyi Fan}\textsuperscript{5},
    \textbf{Chengyi Xing}\textsuperscript{4},
    \textbf{Yansong Tang}\textsuperscript{3},  
    \textbf{He Wang}\textsuperscript{2}, 
    \textbf{Wei-Shi Zheng}$^{1,\dagger}$ \\ 
    \footnotesize 
    \textsuperscript{1} Sun Yat-sen University
    \textsuperscript{2} Peking University 
    \textsuperscript{3} Tsinghua University  \\
    \footnotesize 
    \textsuperscript{4} Stanford University 
    \textsuperscript{5} Jiangxing Intelligence (Guizhou) Technology Inc.\\
    \footnotesize 
    \footnotesize \href{https://frenkielm.github.io/BiDexGrasp.github.io/}{\textcolor{deepPink}{https://frenkielm.github.io/BiDexGrasp.github.io/}} \\
}

\begin{document}
\maketitle
\begin{figure}[ht]
\vspace{-1cm}
\begin{center}
	\includegraphics[width=0.95 \linewidth]{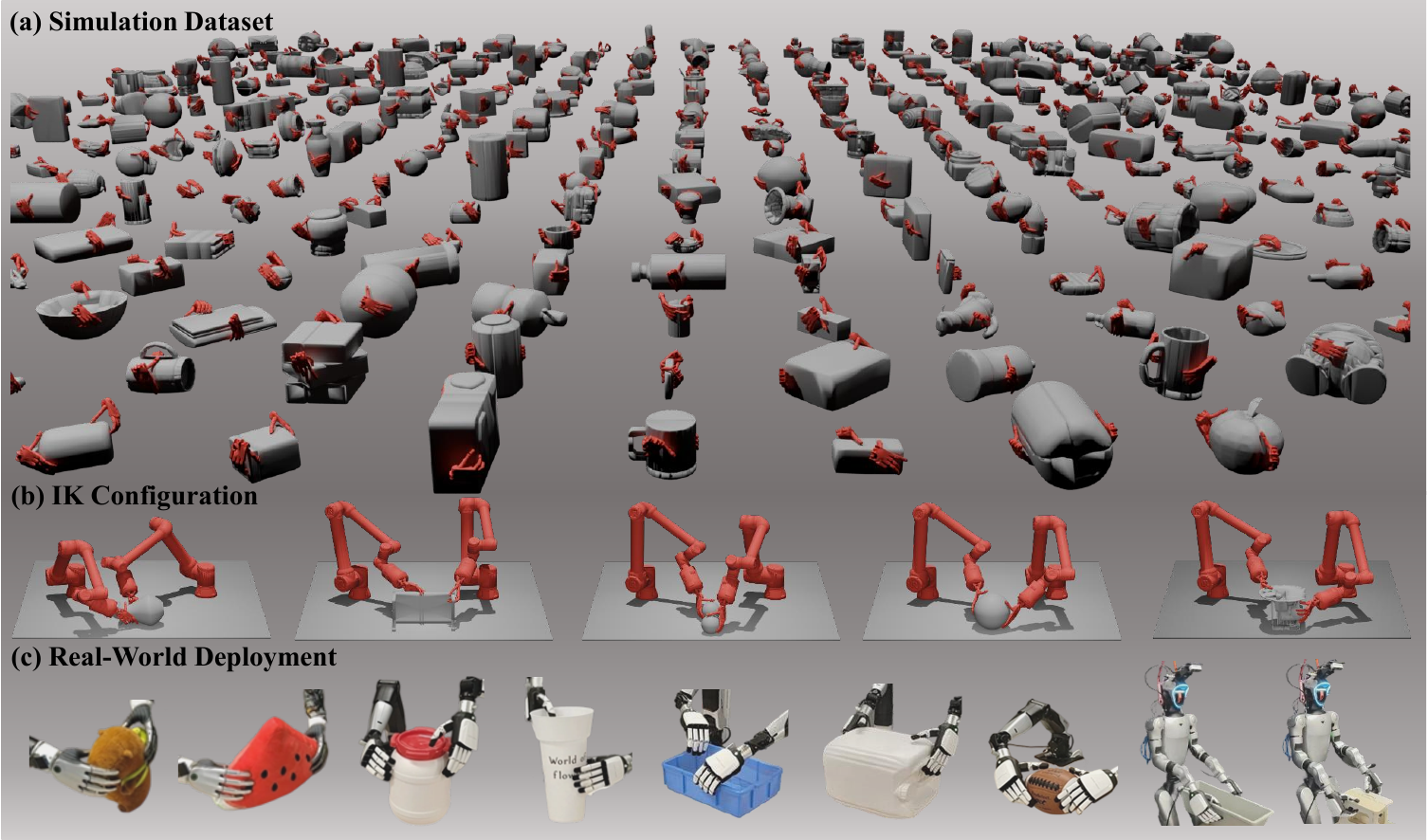}
\end{center}
% \vspace{-7pt}
% \captionsetup{font=footnotesize}
\captionsetup{font=small}
 \caption{Overview of BiDexGrasp. We construct a large-scale, high-quality dataset of coordinated bimanual dexterous grasps with diverse object geometries and sizes, along with feasible IK configurations in tabletop setting. Based on this dataset, we propose a novel data-driven learning-based framework to generate physically feasible dexterous grasps on unseen objects in the real world.}
%   \label{fig:teaser}
% \vspace{-2pt}
\label{fig:first_figure}
\end{figure}

\begin{abstract}
  Bimanual dexterous grasping is a fundamental and promising area in robotics, yet its progress is constrained by the lack of comprehensive datasets and powerful generation models. In this work, we propose BiDexGrasp, consisting of a large-scale bimanual dexterous grasp dataset and a novel learning-based framework. For dataset construction, we propose a novel bimanual grasp synthesis pipeline to efficiently annotate physically feasible data. This pipeline addresses the challenges of high-dimensional bimanual grasping through a two-stage synthesis strategy of efficient region-based grasp initialization and decoupled force-closure grasp optimization. Powered by this pipeline, we construct a large-scale bimanual dexterous grasp dataset, comprising 6351 diverse objects with sizes ranging from 30 to 80 cm, along with 9.53 million annotated grasp data. Based on this dataset, we further introduce a novel learning-based dexterous grasping generation framework. The framework lies in two key designs: a bimanual coordination module and a geometry-size-adaptive grasp generation strategy to generate coordinated and high-quality grasps on unseen objects. Extensive experiments conducted in both simulation and real world demonstrate the superior performance of our proposed data synthesis pipeline and learned generative framework.
\end{abstract}
% \keywords{CoRL, Robots, Learning} 

\section{Introduction}
Robotic dexterous grasping enables human-like object interaction~\cite{zhang2025robustdexgrasp,zhong2025dexgraspvla,cui2025dexgrasp-vla,li2025maniptrans,shaw2024bimanualdexterity,wei2025omnidexgrasp,yuan2025demograsp,lin2025typetele}. Data-driven approaches achieve promising results, showing demands for high-quality datasets and advanced generative models~\cite{fang2020graspnet1billion,wang2022dexgraspnet,zhang2024dexgraspnet2.0,he2025dexvlg,ye2025dex1b}. Most prior studies concentrate on single-hand grasping~\cite{liu2021synthesizing,liu2020synthesizing2, xu2023unidexgrasp,huang2023scenediffuser,shao2020unigrasp}. However, diverse object shapes and sizes restrict single-hand manipulation, making bimanual dexterous grasping indispensable.

Currently, several recent studies explore data synthesis and data-driven methods for bimanual dexterous grasping~\cite{shao2024bimangrasp, li2025dhagrasp} but still restricted to objects of limited size and geometry.
We attribute this to the dual challenges of varied object sizes and the enlarged bimanual search space, which jointly degrade synthesis efficiency and grasp quality. 
To address this, we introduce BiDexGrasp in this work, which contributes a large-scale, high-quality bimanual grasp dataset produced by a novel synthesis pipeline, together with a bimanual-coordinated, geometry- and size-adaptive generation framework that learns from this dataset to produce high-quality grasps on diverse unseen objects.

For dataset construction, we propose an effective bimanual dexterous grasp synthesis pipeline to address the increased complexity and low efficiency caused by the expanded action space. For efficiency, we introduce a bimanual region-constrained grasp initialization strategy to generates coordinated and physically feasible candidates. For quality, we propose a decoupled force-closure grasp optimization strategy that separates the force-closure constraints of each hand, reducing optimization complexity while preserving high-quality per-hand grasps. With this pipeline, we build a large dataset of 9.53M grasps over 6,351 objects spanning diverse geometries and scales (Tab.~\ref{tab:dataset_comparison}).

Built on this dataset, we propose the \textbf{BiDexGrasp} framework for high-quality grasp generation on diverse unseen objects, tackling two challenges: bimanual coordination and generalization across geometries and scales. To enhance coordination, we introduce a bimanual coordination module to predict coordinated grasp views, to guide the grasp poses generation. 
To enhance adaptability, we propose a geometry-size-adaptive grasp strategy that compactifies the action space and emphasizes local structural learning, greatly improving robustness to diverse object geometries and scales.

Extensive simulation and real-world experiments validate both the synthesis pipeline and the generation framework. Our pipeline achieves over 2.8$\times$ higher success rate and 30$\times$ faster synthesis than prior work, and our framework attains 66.8\% success in simulation and 74.6\% on real experiments, demonstrating consistently strong performance across both settings.

\section{Related work}

\subsection{Bimanual Dexterous Grasp}
Bimanual dexterous grasping extends grasping to large and heavy objects but remains underexplored due to the high dimension of the combined degrees of freedom. 
Reinforcement learning approaches~\cite{zhang2024artigrasp, chen2023bi-dexhands} suffer from limited scalability and heavy reliance on hand-crafted rewards, while data-driven methods~\cite{shao2024bimangrasp, li2025dhagrasp, yang2026ultradexgrasp} are constrained by the lack of high-quality datasets and suboptimal model architectures. 
In this work, we address these challenges with a large-scale dataset with high diversity and quality, and a novel generative framework tailored for bimanual dexterous grasping.
% that extend single-hand pipelines to the bimanual setting 
\subsection{Dexterous Grasp Dataset}
The availability of large-scale and high-quality datasets is crucial for the development of robotic~\cite{fang2020graspnet1billion, zhang2024dexgraspnet2.0, ye2025dex1b, mu2024robotwin}. 
Previous research on single-hand dexterous grasping has explored energy-based optimization methods~\cite{chen2025bodex, wang2022dexgraspnet, liu2021synthesizing, liu2020synthesizing2, zurbrugg2025graspqp, chen2025dexonomy}, enabling parallel synthesis of large-scale dexterous grasping data. 
However, in the bimanual setting, the expanded action space and coordination requirements push existing pipelines~\cite{shao2024bimangrasp, li2025dhagrasp, yang2026ultradexgrasp} into low efficiency and over-reliance on single-hand data, yielding datasets of narrow object diversity.
In this work, we introduce a novel bimanual grasp synthesis pipeline and a corresponding large-scale dataset with broad geometric and scale coverage.

\begin{table}[t]
\centering
% \vspace{-0.6cm}
% \captionsetup{font=footnotesize}
\captionsetup{font=small}
\caption{Comparison with previous dexterous grasping datasets}
\label{tab:dataset_comparison}
\scriptsize
\setlength{\tabcolsep}{6pt} % 极紧凑
\renewcommand{\arraystretch}{0.8}

\begin{tabular}{l ccccccccc}
\toprule
\textbf{Dataset} & \textbf{Bim.} & \textbf{Obj} & \textbf{Grasp} & \textbf{Max} & \textbf{Range} & \textbf{Table} & \textbf{IK} & \textbf{Pre-Grasp} & \textbf{Data Independent} \\ 
\midrule
DexGraspNet\cite{wang2022dexgraspnet} & $\times$ & 5355 & 1.32M & 30cm & 28cm & $\times$ & $\times$ & $\times$& \checkmark \\
BODex\cite{chen2025bodex} & $\times$ & 2397 & 3.08M & 24cm & 12cm & \checkmark & \checkmark&  \checkmark & \checkmark \\
DHAGrasp\cite{li2025dhagrasp} & \checkmark & 802 & 1.30M & 30cm & 20cm & $\times$ & $\times$& $\times$ & $\times$ \\
BimanGrasp\cite{shao2024bimangrasp} & \checkmark & 900 & 0.15M & 40cm & 20cm & $\times$ & $\times$& $\times$ & \checkmark \\
\midrule
\textbf{BiDexGrasp} & \checkmark & \textbf{6351} & \textbf{9.53M} & \textbf{80cm} & \textbf{50cm} & \checkmark & \checkmark&  \checkmark & \checkmark \\
\bottomrule
\end{tabular}

\vspace{-0.4cm}
\end{table}

\subsection{Data-Driven Dexterous Grasp Framework}
Data-driven dexterous grasping frameworks have shown strong generalization to unseen objects~\cite{fang2025anydexgrasp, chen2025clutterdexgrasp, huang2025fungrasp}. Most existing methods rely on generative models~\cite{xu2023unidexgrasp, jiang2021grasptta, wei2024d}, and extended through richer conditioning~\cite{wei2025afforddexgrasp}, dedicated training objectives and strategy~\cite{lu2024ugg, wei2024graspasyousay}, physical-guided sampling~\cite{zhong2025dexgraspanything}  or bimanual extensions~\cite{shao2024bimangrasp, shaw2024bimanualdexterity}. However, they limit in the adaptation to diverse object geometries and sizes. 
% In this paper, drawing inspiration from prior methods~\cite{zhang2024dexgraspnet2.0}, our framework explicitly models bimanual coordination and adapts to object geometry and scale, producing robust and physically consistent bimanual grasps.
In this paper, drawing inspiration from prior methods~\cite{zhang2024dexgraspnet2.0}, our framework explicitly models bimanual coordination, producing robust bimanual grasps across object geometry and scale.

\section{Preliminaries}

\textbf{Grasp Wrench Space (GWS).}
Given an object with $m$ contact points, each contact $i$ is characterized by a position $\mathbf{p}_i$, surface normal $\mathbf{n}_i$, and a friction-cone-constrained local force $\mathbf{f}_i$, which together induce a 6D wrench $\mathbf{w}_i = \mathbf{G}_i \mathbf{f}_i$ via the grasp matrix $\mathbf{G}_i$. The set of feasible wrenches at contact $i$ is denoted $\mathcal{W}_i$, and the overall grasp wrench space is the Minkowski sum $\mathcal{W} = \bigoplus_{i=1}^m \mathcal{W}_i$, which characterizes all wrenches the grasp can exert on the object. The minimum distance from the origin to the GWS boundary (GWB) reflects contact stability---a larger distance indicates a more stable grasp.

\textbf{QP-Energy.}
To evaluate the ability of a grasp to resist external disturbances, we adopt a differentiable metric based on quadratic programming. Given six target disturbance wrenches $\{\mathbf{t}_j\}_{j=1}^6$ along the $\pm x, \pm y, \pm z$ axes, the energy is defined as $Q = \sum_{j=1}^{6} \min_{\mathbf{w}_j \in \mathcal{W}} \left\| \beta \mathbf{t}_j - \mathbf{w}_j \right\|^2$, where $\beta$ is a scaling factor. Intuitively, $Q$ measures how well the grasp wrench space $\mathcal{W}$ can approximate disturbance directions. The detailed grasp matrix and QP formulation are provided in Appendix~\ref{sec:appendix_prelim}.

\section{BiDexGrasp Dataset}
\subsection{Overview}
\textbf{Problem Definition.} Given the object mesh $\mathcal{O}$, the goal of grasp data synthesis is to obtain bimanual dexterous grasp poses $\mathcal{G}^{bi} = (\mathcal{G}^{left}, \mathcal{G}^{right}) = ((t^r,r^r,q^r), (t^l,r^l,q^l))$, 
where $r$ and $t$ refer to the rotation and translation of hand wrist, and $q$ refers to the joint poses of dexterous hand.

\begin{figure*}[t]
\centering
\includegraphics[width=1.0\linewidth]{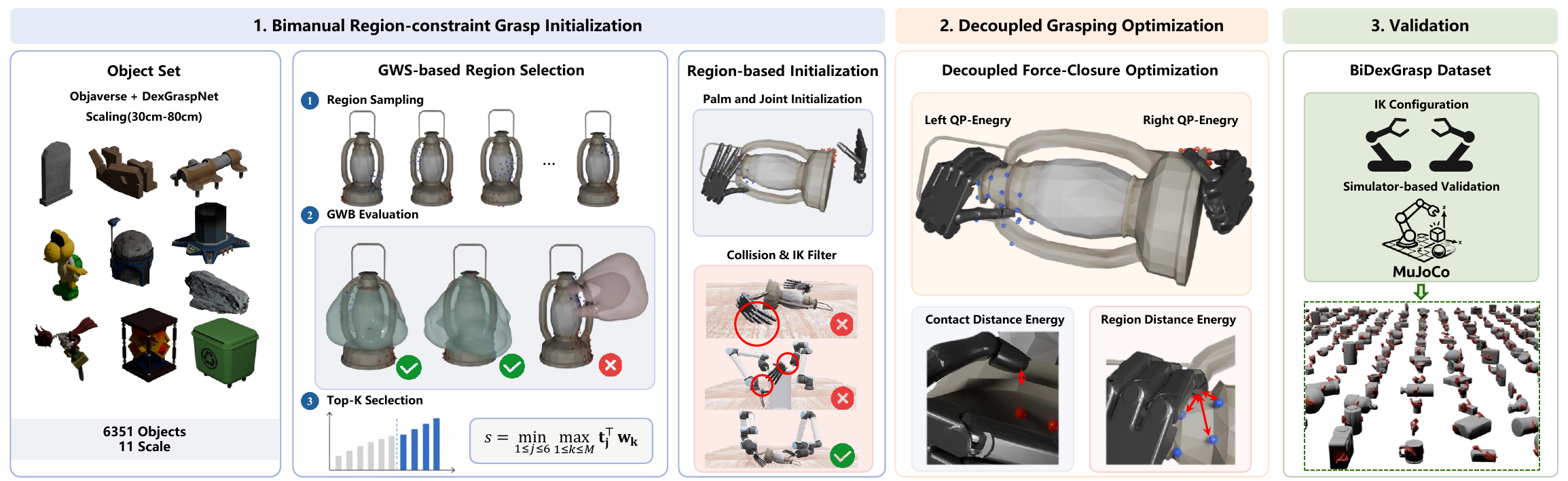}
% \captionsetup{font=footnotesize}
\captionsetup{font=small}
\caption{The data synthesis pipeline for bimanual grasp. (1) Fast constraint-aware bimanual grasp initialization for physically feasible grasp bimanual candidate generation. (2) Decoupled force-closure optimization refines initial candidates into valid bimanual grasps. (3) Arm reachability analysis yields kinematic feasible poses, with MuJoCo~\cite{todorov2012mujoco} simulation filtering invalid grasps.}
\label{fig: pipeline}
\vspace{-0.5cm}
\end{figure*}

\textbf{Synthesis Overview.} We first collect 6,351 objects from~\cite{wang2022dexgraspnet} and ~\cite{deitke2023objaverse} and scale each to 11 discrete sizes from 30 to 80 cm. The synthesis pipeline is consisted of three stages as shown in Figure \ref{fig: pipeline}.

\subsection{Bimanual Region-constraint Grasp Initialization}
\textbf{Challenges for Bimanual Grasp Initialization.}
Naively extending single-hand initialization causes quadratic sampling complexity, and most random pairs violate bimanual constraints. We address this with a bimanual region-constrained initialization strategy: grasping region candidates are first selected via the grasp wrench space (GWS), and initial poses are then sampled within these regions, yielding both efficient and reliable initialization.

\textbf{GWS-based Bimanual Grasping Region Selection.}
This module leverages GWS energy to efficiently filter bimanual grasping regions with the potential for stable, physically feasible grasps.
Specifically, the region selection process consists of 3 steps. (1) We first apply farthest point sampling to select $K_a = 200$ anchors and sample $k = 256$ points within an 8\,cm neighborhood around each anchor to define $K_a$ regions and $K_a^2$ region pairs.
(2) We exclude the pairs with insufficient inter-region distance, and evaluate the remain pairs' stability using the Grasp Wrench Boundary (GWB) estimated by the TDG estimator~\cite{chen2023tdg}.
Specifically, $N = 5$ contact points are sampled per region, and the resulting GWB is represented by a set of boundary wrenches
$\{\mathbf{w}_k\}_{k=1}^{M}$ with $M = 1000$.
To assess robustness against external disturbances, we adopt the six disturbance wrenches
$\{\mathbf{t}_j\}_{j=1}^{6}$ defined in the preliminaries.
The stability score is computed as the minimum projection of the GWB onto these disturbance directions: $s = \min_{1 \le j \le 6} \; \max_{1 \le k \le M} \; \mathbf{t}_j^\top \mathbf{w}_k .$
% \begin{equation}
% s = \min_{1 \le j \le 6} \; \max_{1 \le k \le M} \; \mathbf{t}_j^\top \mathbf{w}_k .
% \end{equation}
(3) Finally, we rank region pairs by stability score, and the top $K_r = 40$ are retained as candidates for grasp initialization.

\textbf{Region-based Grasp Initialization.}
Given the selected regions, we initialize bimanual grasps $\mathcal{G}_{\text{initial}}$ by placing each palm at the closest point on the dilated convex hull to the region center, with orientation aligned to the surface normal. We then enforce collision checking and inverse kinematics (IK) feasibility, discarding invalid configurations before optimization.

\subsection{Decoupled Grasping Optimization}
\textbf{Challenges for Bimanual Grasp Optimization.}
Bimanual grasp optimization is inherently challenging because the enlarged search space induces a tightly coupled objective that destabilizes optimization and often yields imbalanced solutions, where one hand dominates stability while the other contributes marginally, leading to suboptimal local minima.

\textbf{Decoupled Force Closure.}
To obtain high-quality and physically feasible bimanual grasps, we propose a novel optimization strategy which decompose the bimanual force closure energy into single-hand energy terms, allowing each hand to focus on establishing stable local contacts. 
This decoupling reduces optimization complexity and discourages single-hand-dominated solutions, improving both grasp quality and convergence stability.
Specifically, we formulate the overall optimization as follows to optimize $\mathcal{G}^{bi}$ 
by minimizing the QP-Energy decoupled into two single-hand energy terms:
\begin{equation}
\begin{aligned}
\min_{\mathcal{G}^{bi}} 
&  (w_{QP}Q(\mathcal{G}^{left}) + w_{QP}Q(\mathcal{G}^{right}) + w_{dis}E_{dis}+w_{region}E_{region}), \\
\text{s.t.}\quad
& \mathcal{G}^{bi}_{\min} \le \mathcal{G}^{bi} \le \mathcal{G}^{bi}_{\max},\quad \mathbf{c}_{i,w} = \mathrm{FK}(\mathcal{G}^{bi}, \mathbf{c}_{i,l}) , \quad\text{No collision.}
\end{aligned}
\end{equation}

$\mathbf{c}_{i,l}$ and $\mathbf{c}_{i,w}$ denote the positions of the predefined hand contact point $i$ expressed in the local link frame and the world frame. In the QP formulation, the grasp matrix is constructed by associating each $\mathbf{c}_{i,w}$ with its closest point $\mathbf{p}_i$ on the object surface, while feasibility is ensured through joint-limit and collision-avoidance constraints.
We introduce a distance term $E_{dis}$ to encourage proximity between $\mathbf{c}_{i,w}$ and $\mathbf{p}_i$, and a region consistency term $E_{region}$ to keep the contacts close to the initialized contact regions. 
The detailed formulations are provided in Appendix~\ref{sec:appendix_datasyn} and ~\ref{sec:appendix_prelim}.

\subsection{Kinematics-Aware Arm-Hand Configuration Synthesis}
To demonstrate the practical feasibility of the synthesized grasps, our framework supports solving the inverse kinematics (IK) for wrist poses, enabling reachable tabletop configurations. To avoid collisions and facilitate force application, we first synthesize a pre-grasp pose $\mathcal{G}^{bi}_{pre}$, maintaining a contact distance of 1cm, following~\cite{chen2025bodex}. 
And we compute a squeeze pose as the execution target: $\mathcal{G}^{bi}_{squ} = 2 \cdot \mathcal{G}^{bi} - \mathcal{G}^{bi}_{\text{pre}}$. Finally, we generate arm joint trajectories from pre-grasp to squeeze and lift.
\begin{figure*}[]
\centering
% \vspace{-5mm}
\includegraphics[width=1.0\linewidth]{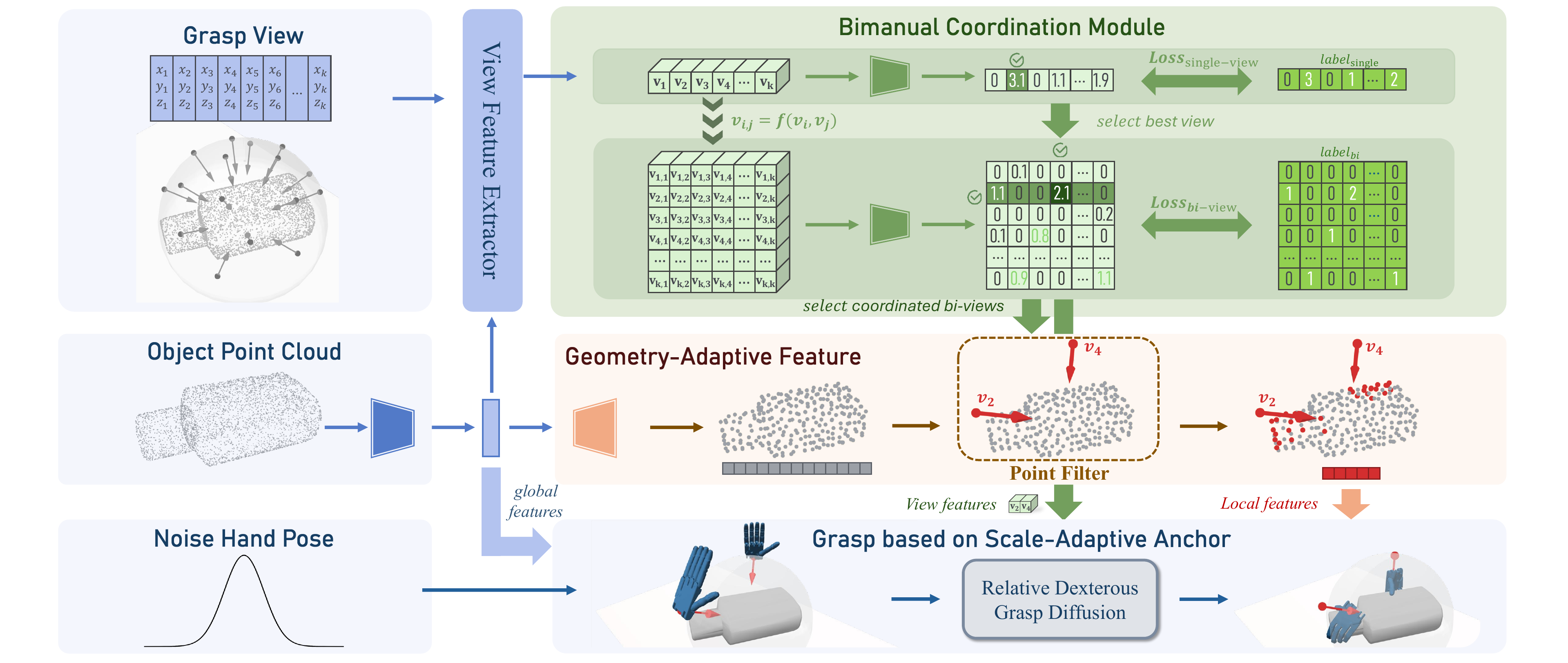}
\captionsetup{font=small}
\caption{ BiDexGrasp Framework. Given the input object point cloud and pre-defined grasp view, the framework first predicts a pair of coordinated grasp views for the two hands by Bimanual Coordination Module. Based on this view, we extract geometry-adaptive object feature and determine the scale-adaptive grasp. Then the relative grasp diffusion is employed to predict relative dexterous grasp form the grasp anchor.
}
\vspace{-4mm}
\label{fig: model}
\end{figure*}

\section{BiDexGrasp Framework}

\textbf{Problem Definition.}
To learn generable bimanual dexterous grasp, we design a data-driven generation model which take object point cloud $\mathcal{O}_{pc}$ as input and generate bimanual dexterous grasp poses $\mathcal{G}^{bi}$.

\textbf{Challenge in Bimanual Grasping.}
Bimanual grasp generation introduces two key challenges. 
(1) \textit{Bimanual coordination}: Bimanual hands must collaboratively target optimal grasp view to ensure global stability while avoiding inter-hand collisions. 
(2) \textit{Adaptation across object geometries and sizes}: The vast diversity in object sizes and geometries greatly hinders efficient model learning, increasing the difficulty of object feature extraction and expanding the action space simultaneously.

\textbf{Model Overview.}
To address these challenges, we propose \textbf{BiDexGrasp}, which achieves bimanual-coordinated and geometry-size-adaptive grasp generation through two components: 
(1) a \emph{bimanual coordination module} that explicitly models inter-hand relationships via coordinated-view prediction; 
(2) a \emph{geometry-size-adaptive generation strategy} that improves cross-object generalization through adaptive grasp anchors. 
The model pipeline is shown in Fig.~\ref{fig: model}.

\subsection{Bimanual Coordination Module.}
To explicitly model bimanual coordination, we first estimate coordination-aware grasping views to guide subsequent grasp prediction. The module first predicts the most promising view as the primary grasping view, and predicts the second view that is most compatible with the primary view. 

Specifically, we uniformly sample $K$ discrete approach directions on the object bounding sphere to construct a candidate view set $\mathcal{V}=\{\mathbf{v}_i\}_{i=1}^{K}$, which formulate view estimation as a view probabilistic regression. To predict the primary view, we introduce $K$ learnable view embeddings $\mathbf{v}^{emb}_{i}$ to extract view features $\mathbf{F}_{\text{view}}=\{\mathbf{f}_i\}_{i=1}^{K}$ by cross-attention: $\mathbf{f}_i = \mathrm{Attn}(\mathbf{v}^{emb}_{i}, \mathbf{F}_{global}, \mathbf{F}_{global})$
, where $\mathbf{F}_{\text{global}}$ are point features extracting by PointNet++~\cite{qi2017pointnet++}. To predict the second view, we model interactions between the primary view and the remaining candidate views to construct bi-view features by a MLPs layer: $\mathbf{f}^{bi}_{i,j} = MLP\!\big(
cat(
\mathbf{f}_{i},
\mathbf{f}_{j},
\mathbf{f}_{i}-\mathbf{f}_{j},
\mathbf{f}_{i}\odot\mathbf{f}_{j}
)\big).$
% \mathbf{v}_{i},
% \mathbf{v}_{j},
% \mathbf{v}_{i}-\mathbf{v}_{j},
% \mathbf{v}_{i}\odot\mathbf{v}_{j}
% )\big).$
These features predict the single-view and bi-view probabilities $p_i$ and $p^{bi}_{i,j}$, supervised by SmoothL1 losses against ground-truth probabilities $y_i$ and $y^{bi}_{i,j}$ derived from the successful-grasp distribution of each object. For single-view, we assign each grasp to the nearest predefined candidate view, then count the number of grasps associated with each view and normalize these counts to form a probability. Similarly, the bi-view ground-truth probabilities are obtained by considering pairs of grasps. Each grasp pair is mapped to the nearest candidate view pair, and the counts are normalized over all candidate pairs to produce a probability ground-truth.

\begin{wrapfigure}{r}{0.37\textwidth}
  \centering
  \vspace{-0.5cm}
  \includegraphics[width=\linewidth]{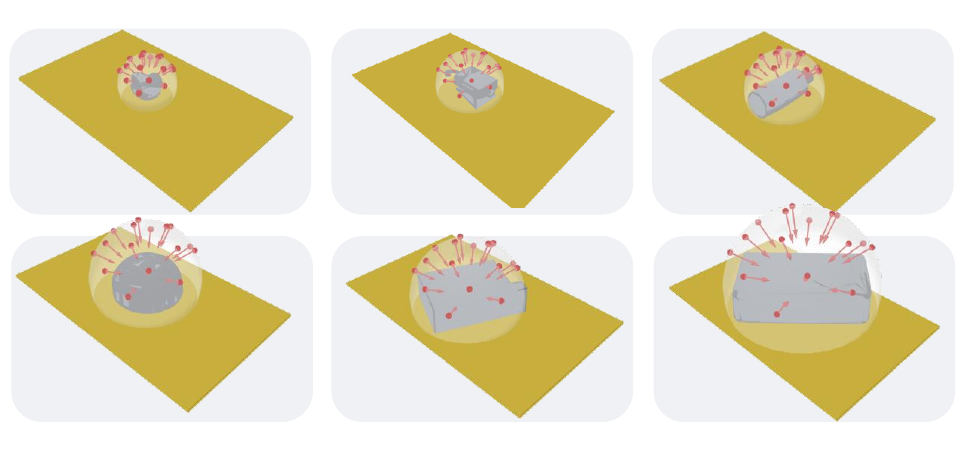}
  % \vspace{-6mm}
% \captionsetup{font=footnotesize}
\captionsetup{font=small}
  \caption{Visualization of grasp anchors.}
  \label{fig:anchor}
  \vspace{-0.2cm}
\end{wrapfigure}

\subsection{Geometry-Size-Adaptive Grasp Generation Strategy}
\textbf{Scale-Adaptive Grasp Anchor}
To improve adaptability to object size, we formulate the grasp wrist prediction as a relative pose $(\Delta t,\Delta r)$ prediction with respect to a grasp anchor $g_{anchor}=(t_{anchor}, r_{ancho})$, which is adapted to the size of the object. 
\begin{align}
t_{pred} &= t_{anchor} + \Delta t, \quad
r_{pred} = r_{anchor} \oplus \Delta r.
\end{align}
The adaptive grasp anchor is determine dynamically according to the grasp view and object point cloud. Specifically, each predicted view intersects with minimum bounding sphere surface, producing an grasp anchor. This anchor-based formulation converts the original absolute pose prediction problem into a relative regression task, significantly reducing the search space of grasps and improve the adaptation to object scale.

\textbf{Geometry-Adaptive Object Feature}
To improve adaptability to object geometry, we extract local geometric features around each predicted grasp anchor, as global features may be dominated by irrelevant structures. Specifically, the global point cloud is first upsampled into $K_p$ representative points using farthest point sampling (FPS). For each predicted grasp anchor, features are then aggregated from its $K_c$-nearest neighbors (KNN). This local feature $\mathbf{F}_{\text{local}}$ captures fine-grained surface geometry around potential contact regions while filtering out distant or grasp-irrelevant structures, which facilitates geometry-adaptive grasp generation. Finally, the local features, view features, and global features are concatenated along the token dimension to form the final object feature representation. 
\begin{equation}
\mathbf{F}_{ga-obj} = \operatorname{Concat}\big(\mathbf{F}_{\text{local}}, \mathbf{F}_{\text{view}}, \mathbf{F}_{\text{global}}\big).
\end{equation}

\begin{table*}[t]
    \centering
% \captionsetup{font=footnotesize}
\captionsetup{font=small}

    \caption{Performance comparison of dual-arm grasping on DexGraspNet and Objaverse datasets.}
    \label{tab:comparison}
    % \small
    \scriptsize

    \renewcommand{\arraystretch}{1.2} 
    \setlength{\tabcolsep}{8pt}
    
    \begin{tabular}{lcccccccc}
        % \toprule
        % \multirow{2}{*}{\textbf{Method}} & \multicolumn{3}{c}{\textbf{SUC $\uparrow$}} & \multirow{2}{*}{\makecell[c]{\textbf{S}\textbf{$\uparrow$}}} & \multirow{2}{*}{\makecell[c]{\textbf{PD}\textbf{$\downarrow$}}} & \multirow{2}{*}{\makecell[c]{\textbf{SPD}\textbf{$\downarrow$}}} & \multirow{2}{*}{\makecell[c]{\textbf{CDC}\textbf{$\downarrow$}}} & \multirow{2}{*}{\makecell[c]{\textbf{D}\textbf{$\downarrow$}}} \\
        % \cmidrule(lr){2-4}
        % & \textbf{SUC-F} & \textbf{SUC-L} & \textbf{SUC-A} & & & & & \\
        \toprule
        \textbf{Method} 
        & \textbf{SUC-F$\uparrow$} 
        & \textbf{SUC-L$\uparrow$} 
        & \textbf{SUC-A$\uparrow$} 
        & \textbf{S$\uparrow$} 
        & \textbf{PD$\downarrow$} 
        & \textbf{SPD$\downarrow$} 
        & \textbf{CDC$\downarrow$} 
        & \textbf{D$\downarrow$} \\
% \midrule
        
        \bottomrule

        \rowcolor{gray!20} 
        \multicolumn{9}{l}{\textit{\textbf{DexGraspNet}}} \\
        % \midrule
        BimanGrasp~\cite{shao2024bimangrasp} & 26.80& -- & -- & 0.16&1.52 &0.26 & 1.87& \textbf{0.150}\\

        % \midrule
        Ours(Float) &75.84 & --& --& 6.73& 0.17&0.02 & 0.59& 0.190\\
        Ours(TableTop) &\textbf{77.11} & \textbf{80.83}&\textbf{71.34} &\textbf{6.87} & \textbf{0.15}&\textbf{0.02} & \textbf{0.52}&0.165 \\
        
        \bottomrule

        \rowcolor{gray!20} 
        \multicolumn{9}{l}{\textit{\textbf{Objaverse}}} \\
        % \midrule
        BimanGrasp~\cite{shao2024bimangrasp} &30.21 & --& --&0.16 &0.87 &0.23 &1.38 & \textbf{0.156}\\

        % \midrule
        Ours(Float) &\textbf{70.82} & --&-- &4.55 &0.20 & \textbf{0.04} &0.67 & 0.180\\
        Ours(TableTop) & 62.41& \textbf{70.41}&\textbf{62.40} & \textbf{5.10}& \textbf{0.20}& 0.05& \textbf{0.67}& 0.158\\
        \bottomrule
    \end{tabular}
\vspace{-0.4cm}
    
\end{table*}

\textbf{Relative Dexterous Grasp Diffusion}
We propose a relative dexterous grasp diffusion to generate bimanual relative wrist poses $(\Delta t^{bi},\Delta r^{bi})$ and dexterous joint poses $q^{bi}$, based on the scale-adaptive grasp anchor $g_{anchor}=(t_{anchor}, r_{anchor})$ and geometry-adaptive object feature $\mathbf{F}_{ga-obj}$. We adopt the DDPM sampling process~\cite{ho2020ddpm}, which can be formalized as follows:
\begin{equation}
\begin{aligned}
p_{\theta}(\mathcal{G}_{0:T}^{\text{bi,rela}} \mid \mathbf{F}_{\text{ga-obj}}) 
&= p(\mathcal{G}^{\text{bi,rela}}_T) \times \prod_{t=1}^T 
p_{\theta}\Big(\mathcal{G}^{\text{bi,rela}}_{t-1} \,\Big|\, 
\mathcal{G}^{\text{bi,rela}}_{t}, \mathbf{F}_{\text{ga-obj}}\Big),
\end{aligned}
\label{eq:grasp_diffusion}
\end{equation}
where $\mathcal{G}^{bi,rela} = (\Delta t^{bi},\Delta r^{bi},q^{bi})$, and $p(\mathcal{G}^{bi,rela})$ is modeled as a Gaussian distribution. During training, we apply L2 losses separately to the translation, rotation, and joint parameter regressions. In addition, we incorporate Chamfer loss~\cite{fan2017chamfer} to explicitly supervise hand geometry. We further introduce a physics-based loss to penalize hand–object penetration as well as intra- and inter-hand self-penetration. All details could be found in Appendix~\ref{sec:appendix_framework}. 
\begin{equation}
\begin{split}
\mathcal{L} &=
\lambda_{\text{single-view}} \mathcal{L}_{\text{single-view}} +
\lambda_{\text{bi-view}}\mathcal{L}_{\text{bi-view}} +
\lambda_{\text{para}}\mathcal{L}_{\text{para}} + \lambda_{\text{chamfer}}\mathcal{L}_{\text{chamfer}} +
\lambda_{\text{physic}}\mathcal{L}_{\text{physic}}.
\end{split}
\end{equation}

\section{Experiments}
\label{sec: experiments}

\subsection{Experiment Settings}

\textbf{Data Synthesis.}
% Data synthesis evaluation is conducted in MuJoCo simulation using the Shadow Hand. The tangential and torsional friction coefficients are set to 0.6 and 0.02, with gravity fixed at $9.8\,\mathrm{m/s^2}$ and object density at $2.5\,\mathrm{kg/m^3}$. We randomly sample 1,000 objects from DexGraspNet and Objaverse, scale each to 6 sizes (30–80 cm), forming 12,000 instances, and synthesize five grasps per instance, resulting in 60,000 grasps per method. The pre-grasp and squeeze poses for BimanGrasp are derived by translating the palm by $\pm 1,\mathrm{cm}$ and perturbing the finger joints by $\pm 0.1\mathrm{rad}$.
Evaluation is conducted in MuJoCo with Shadow Hand, using tangential/torsional friction coefficients of 0.6/0.02, gravity $9.8\,\mathrm{m/s^2}$, object density $2.5\,\mathrm{kg/m^3}$. We randomly sample 1,000 objects from DexGraspNet and Objaverse, scale each to 6 sizes (30--80\,cm) to form 12,000 instances, synthesizing 5 grasps per instance (60,000 grasps per method). For \cite{shao2024bimangrasp}, pre-grasp and squeeze poses are obtained by translating the palm by $\pm 1\,\mathrm{cm}$ and perturbing finger joints by $\pm 0.1\,\mathrm{rad}$.

\textbf{Generation Framework.} The generation framework evaluation is conduct on the DexGraspNet subset of our tabletop dataset. The dataset consists of 2,397 objects, which are randomly split into training and test sets with a 4:1 ratio, resulting in 3,112,117 training grasps. During evaluation, for each test object, we generate 3 grasps for every object pose and scale, yielding a total of 139,956 grasps, which are evaluated under the same evaluation as data generation pipeline.

\subsection{Implements Details}
\textbf{Data Synthesis.} We set $w_{QP}=1000$, $w_{\text{dis}}=100$, $w_{\text{region}}=50$, and generate the dataset on eight NVIDIA 3090 GPUs with 40 parallel worlds per GPU and 20 grasps per world.

\textbf{Generation Framework.}
We set $K=16$, $K_p=256$ and $K_c=16$. For the comparison experiments, we train models for 5 epochs. For the ablation studies, models are trained for 20 epochs on a subset that includes all object scales but only one tabletop pose per object. 
All experiments are implemented in PyTorch and conducted on single RTX 4090 GPU. Additional details are provided in Appendix \ref{sec:appendix_framework}.

\subsection{Evaluation Metrics}
Following~\cite{chen2025bodex}, we evaluate the data synthesis pipeline and generation framework with the following metrics (see Appendix~\ref{appendix:metrics} for full definitions).
\textbf{Grasp Success Rate (SUC)} (\%) measures grasp robustness in simulation under three settings: force-closure against 6 external disturbances (\textbf{SUC-F}), tabletop lifting with a floating hand (\textbf{SUC-L}), and with an arm-controlled hand (\textbf{SUC-A}).
\textbf{Speed (S)} and \textbf{Success Generation Speed (SGS)} (grasps/s) quantify raw and successful synthesis throughput.
\textbf{Penetration Depth (PD)} (cm) and \textbf{Self-Penetration Depth (SPD)} (cm) measure the physical plausibility of hand–object contact and of the hand configuration itself (intra- and inter-hand collisions), respectively.
\textbf{Contact Distance Consistency (CDC)} (cm) characterizes the uniformity of finger contacts, and \textbf{Diversity (D)} (\%) the diversity of the generated grasp distribution.
\begin{table}[t]
    \centering
    \scriptsize
    \setlength{\tabcolsep}{2pt}

    \begin{minipage}[t]{0.335\textwidth}
        \centering
        \captionsetup{font=small}
        \caption{Ablation of data synthesis.}
        \label{tab:ablation_study}
        \begin{tabular*}{\linewidth}{@{\extracolsep{\fill}}lccccc@{}}
            \toprule
            Method & SUC-L & SUC-A & SGS & PD & SPD \\
            \midrule
            Baseline & 28.3 & 12.9 & 2.1 & 0.18 & 0.08 \\
            +R & 44.4 & 39.8 & 2.8 & \textbf{0.13} & 0.03 \\
            +R+Dc & 57.7 & 49.5 & 4.0 & 0.15 & 0.03 \\
            \textbf{+R+Dc+Ps} & \textbf{80.8} & \textbf{71.3} & \textbf{5.6} & 0.15 & \textbf{0.03} \\
            \bottomrule
        \end{tabular*}
    \end{minipage}
    \hfill
    \begin{minipage}[t]{0.305\textwidth}
        \centering
        \captionsetup{font=small}
        \caption{Comparison with SOTAs.}
        \label{tab:sota_comparison}
        \renewcommand{\arraystretch}{1.08}   % <-- 加这一行
        \begin{tabular*}{\linewidth}{@{\extracolsep{\fill}}lccccc@{}}
            \toprule
            Method & SUC-L & PD & SPD & CDC & D \\
            \midrule
            \cite{huang2023scenediffuser} & 15.2 & 3.29 & 0.11 & 3.79 & \textbf{0.174 }\\
            \cite{xu2024dgtr} & 41.5 & 2.24 & 0.13 & 3.02 & 0.201 \\
            \midrule
            \textbf{Ours} & \textbf{66.8} & \textbf{1.53} & \textbf{0.01} & \textbf{2.32} & 0.197 \\
            \bottomrule
        \end{tabular*}
    \end{minipage}
    \hfill
    \begin{minipage}[t]{0.335\textwidth}
        \centering
        \captionsetup{font=small}
        \caption{Ablation of framework.}
        \label{tab:grasp_comparison}
        \begin{tabular*}{\linewidth}{@{\extracolsep{\fill}}ccc|cccc@{}}
            \toprule
            BCM & GAF & SAGA & SUC-L & PD & SPD & CDC \\
            \midrule
            $\times$ & $\times$ & $\times$ & 41.2 & 2.25 & \textbf{0.01} & 2.87 \\
            $\times$ & $\checkmark$ & $\checkmark$ & 44.7 & 2.28 & \textbf{0.01} & 2.96 \\
            $\checkmark$ & $\checkmark$ & $\times$ & 52.7 & 2.05 & 0.02 & 2.82 \\
            $\checkmark$ & $\checkmark$ & $\checkmark$ & \textbf{61.9} & \textbf{1.73} & 0.02 & \textbf{2.56} \\
            \bottomrule
        \end{tabular*}
    \end{minipage}
    \vspace{-0.2cm}
\end{table}
\subsection{Data Synthesis Experiments}

\textbf{Comparison Results.} 
We compare with BimanGrasp, a representative open-source bimanual grasp synthesis pipeline. Since BimanGrasp is restricted to floating-base grasp generation, we evaluate our method in both floating and tabletop settings for a fair comparison. Tab.~\ref{tab:comparison} shows that our pipeline significantly outperforms prior work in both quality (SUC, PD, SPD, CDC) and speed (S). On DexGraspNet, we increase the SUC by roughly $2.8\times$ under both floating-hand and tabletop settings, with a $40\times$ synthesis speedup. On Objaverse, which features more complex geometries, our method maintains strong success rates and high synthesis efficiency.

\textbf{Ablation Study.} Tab.~\ref{tab:ablation_study} shows the ablation study for each component of our synthesis pipeline. The baseline is a straightforward extension of BODex from the single-hand setting to a bimanual setup. The results show that R (Region Initialization), Dc (Decoupled Force-Closure) and Ps (Pre-grasp Strategies) all contribute to improving grasp success rate (SUC-L and -A) and synthesis speed (SGS).

\subsection{Generation Framework Experiments}

\textbf{Comparison Results.} Tab.~\ref{tab:sota_comparison} reports our method against recent SOTA approaches. Our method achieves the highest SUC with the lowest PD and SPD, indicating precise and stable grasps. While diversity is slightly lower, this trade-off is acceptable since our task prioritizes grasp success. Overall, our method outperforms prior methods in both feasibility and reliability.

\textbf{Ablation Study.} Tab.~\ref{tab:grasp_comparison} evaluates each component over a DDPM-based bimanual baseline. The Bimanual Coordination Module (BCM) substantially raises SUC, confirming its role in coordinating dual-hand grasps. The Geometry-Adaptive Feature (GAF) and Scale-Adaptive Grasp Anchor (SAGA) adapt to local geometry and object scale, jointly boosting SUC and reducing SPD and CDC.

\begin{figure}[!t]
\centering
% \vspace{-0.4cm}
\includegraphics[width=1.0 \linewidth]{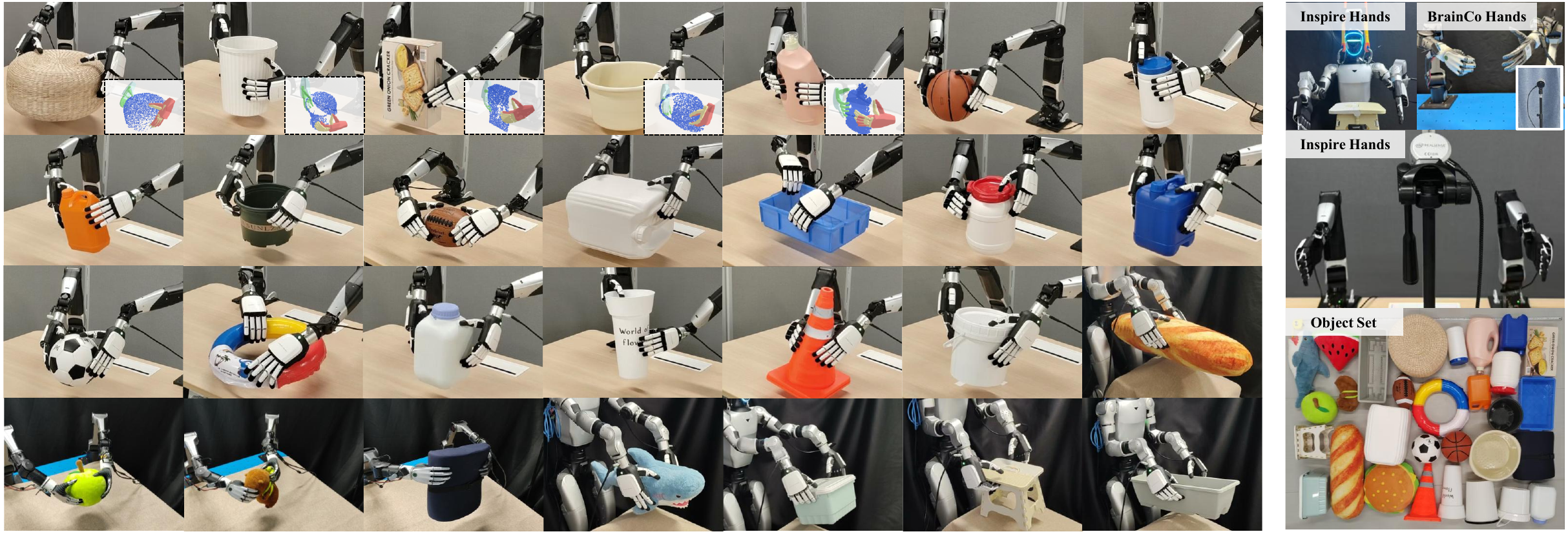}
\vspace{-0.4cm}
% \captionsetup{font=footnotesize}
\captionsetup{font=small}

% \caption{The visualization of our framework in the real world experiments.
\caption{Real-world experiments. Left: qualitative grasp results. Right: hardware setup and object set.
\vspace{-0.4cm}
}
\label{fig:real_result}
\end{figure}
\subsection{Real-World Experiments}
We conduct real-world experiments with two dexterous hands (Inspire and BrainCo) and three robotic arms (Unitree G1, Piper, and Nero), performing 260 trials on 30 objects (setup and qualitative results in Fig.~\ref{fig:real_result}). We evaluate two input settings: single-view and full point clouds. For single-view, we extract object point clouds from RGB-D images via a segmentation model~\cite{ren2024groundedsam}. For full point clouds, we reconstruct object meshes~\cite{hyper3d2024}, estimate 6D poses~\cite{wen2024foundationpose}, and sample the point cloud. ShadowHand grasps are retargeted~\cite{wei2024graspasyousay} to the Inspire and BrainCo hands. Our method achieves 66.0\% success on single-view and 76.7\% on full point clouds, averaging 74.6\%.

\section{Conclusion}
\label{sec:conclusion}
We present BiDexGrasp, a comprehensive solution that advances bimanual dexterous grasping through both a large-scale dataset and a learning-based framework. Our synthesis pipeline yields a dataset of 6,351 diverse objects with 9.53M annotated grasps, on which our generation framework produces high-quality grasps for unseen objects. Extensive experiments in simulation and the real world show that our pipeline significantly improves synthesis efficiency and quality, and our framework generates coordinated, high-quality grasps across diverse objects.

\section{Limitations}
\label{sec:limitations}
First, our synthesis pipeline assumes fixed hand contact points, which restricts the contact-space diversity of generated grasps. 
Second, our method is stability-oriented and may struggle on tasks requiring specific functional contacts; integrating semantic priors into the bimanual grasping region is a promising direction. 
Finally, our model operates in the floating-hand setting and does not use the dataset's IK configurations; joint arm-hand prediction is a natural extension for future work.
\bibliography{example}  % .bib
\clearpage
\appendix
\section{Preliminaries}
\label{sec:appendix_prelim}
\textbf{Contact Model} characterizes the resultant wrench exerted by a grasp on an object. Considering an object $O$ with $m$ contact points, the local contact mechanics at each point $i \in \{1, \dots, m\}$ are modeled as follows:
\begin{equation}
\begin{aligned}
\mathcal{F}_i &= \{ \mathbf{f}_i \in \mathbb{R}^3 \mid 0 \leq f_{i,1} \leq 1, \sqrt{f_{i,2}^2 + f_{i,3}^2} \leq \mu f_{i,1} \}, \\
\mathbf{G}_i &= 
\begin{bmatrix}
\mathbf{n}_i & \mathbf{d}_i & \mathbf{e}_i \\
\mathbf{p}_i \times \mathbf{n}_i & \mathbf{p}_i \times \mathbf{d}_i & \mathbf{p}_i \times \mathbf{e}_i
\end{bmatrix}
\in \mathbb{R}^{6 \times 3},
\end{aligned}
\end{equation}
where $\mathbf{p}_i \in \mathbb{R}^3$ denotes the contact position and $\mathbf{n}_i \in \mathbb{R}^3$ represents the inward-pointing unit surface normal. The vectors $\mathbf{d}_i, \mathbf{e}_i \in \mathbb{R}^3$ are orthogonal unit tangent vectors such that $\mathbf{n}_i = \mathbf{d}_i \times \mathbf{e}_i$, and $\mu$ is the friction coefficient. 

For each contact point, $\mathcal{F}_i$ represents the discretized friction cone and $G_i$ is the grasp matrix. The local contact force $\mathbf{f}_i \in \mathcal{F}_i$ is mapped to a 6D wrench $\mathbf{w}_i \in \mathbb{R}^6$ in the object's coordinate frame via $\mathbf{w}_i=\mathbf{G}_i\mathbf{f}_i$ and the total wrench $\mathbf{w}$ generated by the grasp is the sum of the individual wrenches from all $m$ contact points, i.e., $\mathbf{w} = \sum_{i=1}^m \mathbf{w}_i$.

\textbf{Grasp Wrench Space (GWS)}, denoted as $\mathcal{W}$, encompasses the set of all possible resultant wrenches that can be exerted on the object. For each contact point $i$, we define the individual wrench space as $\mathcal{W}_i = \{ \mathbf{G}_i \mathbf{f}_i\mid\mathbf{f}_i\in\mathcal{F}_i\}$. The aggregate GWS is then constructed via the Minkowski sum of these per-contact wrench spaces $\mathcal{W} = \bigoplus_{i=1}^m \mathcal{W}_i$.

\textbf{QP-Energy} is a differentiable metric designed to evaluate the capability of the current contact state to resist external disturbances $\{\mathbf{t}_j\}_{j=1}^6$ along six orthogonal axes, e.g., $[\pm 1, 0, 0, 0, 0, 0]^\top$. For a fixed contact system $\{\mathbf{G}_i\}$ and disturbances $\{t_j\}$, the QP-Energy $Q$ is calculated as:

\begin{equation}
\begin{aligned}
Q &\triangleq \sum_{j=1}^{6}\min_{\mathbf{f}_{j,1}, ..., \mathbf{f}_{j,m}} \left\| \beta \mathbf{t}_j - \sum_{i=1}^m \mathbf{G}_{i} \mathbf{f}_{j,i} \right\|^2 ,\\
&\text{s.t.}  \mathbf{f}_{j,i} \in \mathcal{F}_i,
\quad \sum_{i=1}^m f_{i,1} \geq \gamma,
\end{aligned}
\label{eq:new}
\end{equation}
where $\beta$ and $\gamma$ are two positive hyperparameters. 
\section{Data Synthesis Details}
\label{sec:appendix_datasyn}
\subsection{Details of Bimanual Region-constraint Grasp Initialization}
\textbf{GWS-based Bimanual Grasping Region Selection} As described in the main paper, we first sample candidate regions on the object surface and generate local regions around them. These regions are then filtered to form valid region pairs, which are subsequently evaluated using the GWB metric, from which the top-performing pairs are selected. This section provides additional details on the region filtering process.

To avoid physically infeasible or redundant grasps, region pairs that are too close are discarded. Given two regions $\mathcal{R}_a$ and $\mathcal{R}_b$, we compute the minimum inter-region distance
\begin{equation}
d(\mathcal{R}_a, \mathcal{R}_b) = \min_{\mathbf{p} \in \mathcal{R}_a,\, \mathbf{q} \in \mathcal{R}_b} \lVert \mathbf{p} - \mathbf{q} \rVert,
\end{equation}
and retain the pair only if $d(\mathcal{R}_a, \mathcal{R}_b) \ge \tau_2$, where $\tau_2 = 5\,\mathrm{cm}$.

\textbf{Region-based Grasp Initialization} 
This section provides additional details on the initialization of hand poses, including the computation of the initial rotation and the initialization of joint angles.

As described in the main paper, for each hand we first select the point on the object’s inflated closure that is closest to the chosen region. The palm center is positioned to face this point, and the palm normal is aligned with the inward-facing normal of the closure at that location. Subsequently, the hand base is rotated around the palm normal to the orientation that is closest to a predefined preferred direction, which is chosen to be kinematically favorable for the robot arm. Specifically, the preferred direction is set to $(1, -1, -1)$ for the left hand and $(1, 1, -1)$ for the right hand, corresponding to forward–downward–right and forward–downward–left orientations, respectively.

For joint angle initialization, we adapt the finger configuration based on whether the associated contact region is locally flat. Flatness is determined by the dispersion of surface normal directions within the region. Given a set of surface normal vectors $\{\mathbf{n}_i\}_{i=1}^{N}$ for a region, we first normalize them and compute the resultant length
\begin{equation}
R = \left\lVert \frac{1}{N} \sum_{i=1}^{N} \frac{\mathbf{n}_i}{\lVert \mathbf{n}_i \rVert} \right\rVert .
\end{equation}
Following directional statistics, we define the angular variance as
\begin{equation}
V = 2 (1 - R).
\end{equation}
A region is classified as flat if $V < \tau$, where $\tau = 0.005$ in all experiments. If the region is flat, the fingers are initialized in a more open configuration; otherwise, they are initialized with a slightly inward-curved posture. 
% Representative initial hand configurations are illustrated in Fig\ref{fig1}.

\subsection{Details of Decoupled Grasping Optimization}
% \subsubsection{Grasping Optimization Energy}

As stated in the main paper, the overall optimization process can be formulated as
\begin{equation}
\begin{aligned}
\min_{\mathcal{G}^{bi}} 
&  (w_{QP}Q(\mathcal{G}^{left}) + w_{QP}Q(\mathcal{G}^{right}) +w_{dis}E_{dis}+w_{region}E_{region}), \\
\text{s.t.}\quad
& \mathcal{G}^{bi}_{\min} \le \mathcal{G}^{bi} \le \mathcal{G}^{bi}_{\max}, \quad
\mathbf{c}_{i,w} = \mathrm{FK}(\mathcal{G}^{bi}, \mathbf{c}_{i,l}), \quad
\text{No collision.}
\end{aligned}
\end{equation}
And we define $E_{dis}$ to encourage the contact points on the hands to be close to the corresponding points on the object surface:
\begin{equation}
E_{dis} = \sum_{i=1}^{m}(||\mathbf{c}_{i,w}-\mathbf{p}_i||)^2.
\end{equation}
We further introduce a region consistency term $E_{region}$ to keep the contact points close to the contact regions identified during initialization, thereby promoting coordinated bimanual grasping:
\begin{equation}
E_{region} = \sum_{i=1}^{m} \phi(d_i),
\end{equation}
where the effective distance $d_i$ from the $i$-th hand contact point to the contact region $\mathcal{R}$ is defined as
\begin{equation}
d_i = \! \min_{\mathbf{r} \in \mathcal{R}} 
\lVert \mathbf{c}_{i,w} - \mathbf{r} \rVert,\; 
\end{equation}
and $\phi(\cdot)$ is a piecewise penalty function given by
\begin{equation}
\phi(d) =
\begin{cases}
0, & d \le a, \\[4pt]
H\!\left(\dfrac{d-a}{b-a}\right)(d-a), & a < d \le b, \\[8pt]
d-a, & d > b,
\end{cases}
\end{equation}
with $a$ denoting a distance threshold, $b=2a$ defining the transition bandwidth, and $H(t)=3t^2-2t^3$ being a cubic Hermite interpolation function.

This objective naturally leads to a bi-level optimization procedure. In the low-level optimization, we fix the grasp parameters $\mathcal{G}^{left}$ and $\mathcal{G}^{right}$, and solve for the grasp quality terms $Q(\mathcal{G}^{left})$ and $Q(\mathcal{G}^{right})$. This step is implemented using ReLU-QP\cite{bishop2024relu}, a PyTorch-based ADMM solver, which enables efficient and parallel computation of QP-based grasp energies across different objects.

In the upper-level optimization, the grasp parameters $\mathcal{G}^{left}$ and $\mathcal{G}^{right}$ are optimized to minimize the weighted sum of all energy terms while satisfying the kinematic and collision constraints. Specifically, we optimize the grasp parameters under joint limits, object contact constraints, and collision-free conditions. To enforce these constraints, we directly adopt the energy functions provided by cuRobo\cite{sundaralingam2023curobo}, including joint limitation energy, self-penetration energy, and inter-hand penetration energy.

\section{Framework Details}
\label{sec:appendix_framework}
\subsection{Model Details}
\subsubsection{Grasp View}
% 我们在一个单位球的上表面均匀地采样K个点，使用这固定的K个点作为抓取的view
We uniformly sample $K$ grasp views on the upper hemisphere of a unit sphere and use these fixed directions as candidate grasp views. Specifically, each grasp view is represented as a unit vector $\mathbf{v}_i \in \mathbb{R}^3$ pointing from the hemisphere surface to the object center. 

\subsubsection{Scale-Adaptive Grasp Anchor}
Given a selected grasp view $\mathbf{v}$, we construct a scale-adaptive grasp anchor $g_{anchor} = (t_{anchor}, r_{anchor})$. The translation component $t_{anchor}$ is determined as the intersection point between the grasp view direction and the object’s minimum bounding sphere, ensuring that the anchor location adapts to the object scale. 

The rotation component $r_{anchor}$ is defined by aligning the $x$-axis of the grasp frame with the direction pointing toward the object center. The $z$-axis is computed as the projection of the world $z$-axis onto the plane orthogonal to the $x$-axis, and the $y$-axis is obtained accordingly to form a right-handed coordinate frame. This construction yields a unique and consistent grasp anchor frame for each grasp view.

\begin{figure}[t]
\centering
\includegraphics[width=\linewidth]{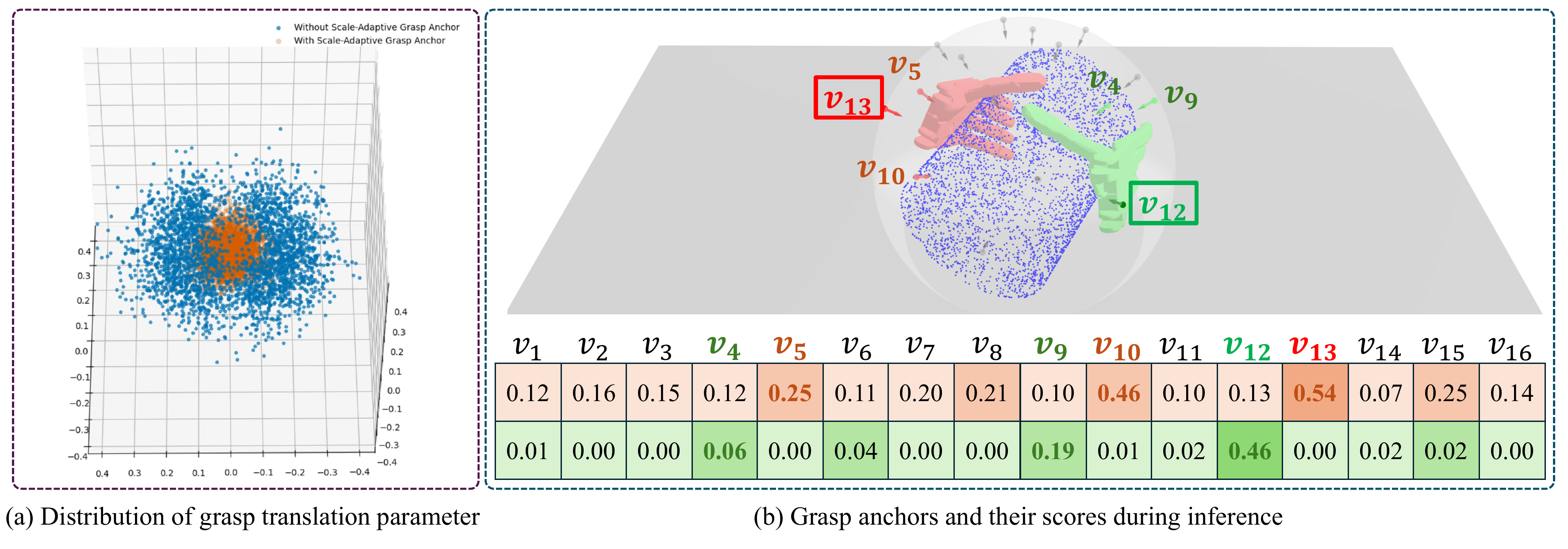}
% \vspace{-0.5cm}
\caption{Visualization of scale-adaptive grasp anchors. 
(a) Distribution of grasp translation parameters represented in the world coordinate system (blue) and in the anchor coordinate system (orange), respectively. Results show the proposed anchor-based formulation effectively reduces the action space for a more compact grasp representation.
(b) Grasp anchors and corresponding view scores during inference, illustrating the conditional view selection process. The model first samples the primary grasp view from the top candidate views, then samples the secondary grasp view via the conditional bi-view distribution.}

\label{fig: graspanchor}
\end{figure}

With the grasp anchor defined, a grasp pose $(t, r)$ expressed in the world coordinate system can be transformed into a relative representation $(\Delta t, \Delta r)$ with respect to the anchor frame. Fig.~\ref{fig: graspanchor}(a) visualizes the distribution of grasp translation parameters represented in the world coordinate system and the anchor coordinate system, respectively. The results show that the proposed anchor-based formulation effectively reduces the action space, leading to a more compact and structured grasp representation.

\subsubsection{Selection of Grasp View in Training and Inference Time}
During training, we adopt a teacher-forcing strategy, where the ground-truth grasp views are directly used to construct the scale-adaptive grasp anchors. This stabilizes training and allows the model to focus on learning grasp generation conditioned on correct view information.

During inference, the model first predicts the single-view probabilistic $\{p_i\} \in \mathbb{R}^K$ and the bi-view probabilistic $\{p^{bi}_{i,j}\} \in \mathbb{R}^{K \times K}$. We select the top-$k$ views with the highest single-view probabilities (with $k=5$), apply temperature scaling with $\tau = 0.1$, and perform softmax sampling to obtain the primary grasp view index $i$. Conditioned on the selected primary view, we extract the corresponding row $\{p^{bi}_{i,j}\}$ from the bi-view probabilistic and apply the same top-$k$ sampling strategy to obtain the second grasp view index $j$.

The overall sampling process is illustrated in Fig.~\ref{fig: graspanchor}(b). In the example, the model first samples the primary grasp view from the top-scoring candidate views $\{5, 8, 10, 13, 15\}$ and selects view $13$. Subsequently, the conditional bi-view distribution $\{p^{bi}_{13,j}\}$ is used to sample the second grasp view from $\{4, 6, 9, 12, 15\}$, resulting in the selection of view $12$.

\subsection{Loss Function Details} 
We provide a detailed description of the loss functions used to train our framework. The overall training objective is defined as:
\begin{equation}
\begin{split}
\mathcal{L} &=
\lambda_{\text{single-view}} \mathcal{L}_{\text{single-view}} +
\lambda_{\text{bi-view}}\mathcal{L}_{\text{bi-view}} +
\lambda_{\text{para}}\mathcal{L}_{\text{para}} \\
&\quad + \lambda_{\text{chamfer}}\mathcal{L}_{\text{chamfer}} +
\lambda_{\text{physic}}\mathcal{L}_{\text{physic}}.
\end{split}
\end{equation}
where the details are as following:

\paragraph{Single-view Loss.}
We supervise the predicted primary grasping view scores using a masked regression loss. Let $p_i$ denote the predicted score for view $\mathbf{v}_i \in \mathcal{V}$ and $y_i$ the ground-truth signal. The loss is defined as
\begin{equation}
\mathcal{L}_{\text{single-view}} =
\frac{\sum_{i=1}^{K} m_i \cdot \mathrm{SmoothL1}(p_i, y_i)}
{\sum_{i=1}^{K} m_i + \epsilon},
\end{equation}
where $m_i \in \{0,1\}$ is a validity mask indicating whether view $i$ has supervision.

\paragraph{Bi-view Loss.}
We supervise the predicted bi-view compatibility using a masked pairwise regression loss. Let $p^{bi}_{i,j}$ denote the predicted compatibility score for view pair $(\mathbf{v}_i, \mathbf{v}_j)$ and $y^{bi}_{i,j}$ the corresponding ground-truth signal. The loss is defined as
\begin{equation}
\mathcal{L}_{\text{bi-view}} =
\frac{\sum_{i=1}^{K} \sum_{j=1}^{K}
m_i \cdot \mathrm{SmoothL1}(p^{bi}_{i,j}, y^{bi}_{i,j})}
{\sum_{i=1}^{K} m_i \cdot K + \epsilon},
\end{equation}
where $m_i \in \{0,1\}$ is the validity mask applied to each row.

\paragraph{Parameter Loss.}
We supervise the predicted bimanual dexterous grasp poses by minimizing the L2 distance between the predicted grasp parameters $\mathcal{G}^{bi}_{pred}$ and the ground-truth grasp parameters $\mathcal{G}^{bi}_{gt}$:
\begin{equation}
\mathcal{L}_{para} 
= \left\| \mathcal{G}^{bi}_{pred} - \mathcal{G}^{bi}_{gt} \right\|_2^2 .
\end{equation}

\paragraph{Chamfer Loss.}
We further impose a geometric consistency loss in 3D space. The predicted grasp parameters $\mathcal{G}^{bi}_{pred}$ are converted into hand meshes via the dexterous hand model, from which surface point clouds $\mathcal{P}_{pred}$ are sampled. Ground-truth point clouds $\mathcal{P}_{gt}$ are obtained in the same way from $\mathcal{G}^{bi}_{gt}$. The Chamfer loss is defined as
\begin{equation}
\mathcal{L}_{chamfer} =
\frac{1}{|\mathcal{P}_{pred}|} \sum_{\mathbf{x} \in \mathcal{P}_{pred}} 
\min_{\mathbf{y} \in \mathcal{P}_{gt}} \|\mathbf{x} - \mathbf{y}\|_2^2
+
\frac{1}{|\mathcal{P}_{gt}|} \sum_{\mathbf{y} \in \mathcal{P}_{gt}} 
\min_{\mathbf{x} \in \mathcal{P}_{pred}} \|\mathbf{y} - \mathbf{x}\|_2^2 .
\end{equation}

\paragraph{Physics-base Losses.}
To ensure physical plausibility of the predicted grasps, we further introduce Physics-base losses to improve physical feasibility. The physics-based losses are:
\begin{equation}
    \mathcal{L}_{\text{physic}} = \lambda_{\text{self-pen}}\mathcal{L}_{\text{self-pen}} + \lambda_{\text{obj-pen}}\mathcal{L}_{\text{obj-pen}} +
    \lambda_{\text{spf}}\mathcal{L}_{\text{spf}}
\end{equation}

\paragraph{Self-bi-penetration Loss.}
We penalize self-collisions of inter-hands and intra-hands. Let $\mathcal{K} = \{\mathbf{k}_i\}_{i=1}^{N}$ denote a set of predefined hand keypoints used for penetration avoidance (including both hands). We compute all pairwise distances and apply a soft penalty when the distance is smaller than a safety threshold $\delta$:
\begin{equation}
\mathcal{L}_{self-pen} =
\sum_{i \neq j}
\max\!\big(0,\; \delta - \|\mathbf{k}_i - \mathbf{k}_j\|_2 \big).
\end{equation}
This term discourages physically infeasible interpenetration between fingers and hands.

\paragraph{Hand–Object Penetration Loss.}
To prevent the hand from penetrating into the object, we compute the signed distance $s_i$ from each hand surface keypoint $\mathbf{p}_i$ to the object surface, where $s_i>0$ indicates the point lies inside the object. The loss is defined as
\begin{equation}
\mathcal{L}_{obj-pen} = \sum_i \max(0, s_i).
\end{equation}

\paragraph{Contact Loss.}
To encourage hand–object contact, we pull predefined hand contact candidate points $\mathcal{C}=\{\mathbf{c}_i\}$ toward the object surface point cloud $\mathcal{O}$. For each point, we compute the nearest-neighbor distance
\begin{equation}
d_i = \min_{\mathbf{o} \in \mathcal{O}} \|\mathbf{c}_i - \mathbf{o}\|_2 .
\end{equation}
Only points within a threshold $\tau$ are considered, and the loss is
\begin{equation}
\mathcal{L}_{spf} =
\frac{1}{|\mathcal{I}|} \sum_{i \in \mathcal{I}} d_i,
\quad
\mathcal{I} = \{ i \mid d_i < \tau \}.
\end{equation}

% \lambda_{\text{single-view}} 10.0
% \lambda_{\text{bi-view}} 1.0
% \lambda_{\text{para}} 15.0
% \lambda_{\text{chamfer}} 0.25
% \lambda_{\text{physic}} 1.0

% \lambda_{\text{obj-pen}} 100.0
% \lambda_{\text{self-pen}} 50.0
% \lambda_{\text{spf}} 50.0

The weighting coefficients of the loss terms are set as follows:
$\lambda_{\text{single-view}} = 10.0$, 
$\lambda_{\text{bi-view}} = 1.0$, 
$\lambda_{\text{para}} = 15.0$, 
$\lambda_{\text{chamfer}} = 0.25$, 
$\lambda_{\text{physic}} = 1.0$, 
$\lambda_{\text{obj-pen}} = 100.0$, 
$\lambda_{\text{self-pen}} = 50.0$, 
and $\lambda_{\text{spf}} = 50.0$.

\section{Evaluation Metric Details}
\label{appendix:metrics}

\textbf{Grasp Success Rate (SUC).} The hand executes a grasp from the pre-grasp pose $\mathcal{G}^{bi}_{pre}$ to the squeeze pose $\mathcal{G}^{bi}_{squ}$. A grasp is considered successful if the object's translation and rotation remain within 5 cm and 15° for at least 3 seconds after execution. \textbf{SUC-F} evaluates force-closure success under external disturbances along six orthogonal directions. \textbf{SUC-L} evaluates lift success in a tabletop scenario with a floating hand. \textbf{SUC-A} evaluates lift success in a tabletop scenario with an arm-controlled hand.

\textbf{Speed (S) and Success Generation Speed (SGS).} These measure the number of raw grasps and successful grasps synthesized per second, respectively. All numbers are obtained on a single NVIDIA GeForce RTX 3090 GPU, while baseline speeds are cited from their original reports.

\textbf{Penetration Depth (PD).} The maximum intersection distance between the object mesh and the hand mesh.

\textbf{Self-Penetration Depth (SPD).} The maximum penetration distance among all hand collision meshes, including both intra-hand and inter-hand self-penetrations.

\textbf{Contact Distance Consistency (CDC).} The range between the maximum and minimum signed contact distances across all fingers, reflecting the uniformity of finger-object contacts.

\textbf{Diversity (D).} The explained variance ratio of the first principal component from PCA applied to the generated grasp distribution, quantifying the concentration (and inversely, diversity) of bimanual grasps.

\section{Experiments of Data Synthesis}

% \subsection{Details of Experiments}

\subsection{Qualitative Experiments of Region Selection}
To validate the effectiveness of our region filtering strategy, we compare our method with random initialization. We randomly select 2,000 objects from DexGraspNet\cite{wang2022dexgraspnet} and Objaverse\cite{deitke2023objaverse}, respectively. For each object, five region pairs are generated using our region generation method and random sampling, and five contact points are sampled within each region to compute the Q1 metric, which measures the distance between the Grasp Wrench Space (GWS) formed by the contact points and the origin, where a larger value indicates a more stable grasp. As shown in Table~\ref{tab:inference_time}, our method consistently achieves higher Q1 values than random sampling on both datasets, demonstrating that the proposed region filtering strategy is more likely to produce stable grasp configurations and generalizes well across both structured and diverse object geometries.

\begin{table}[htbp]
\centering
\begin{tabular}{lcccc}
\toprule
Method & ours(DGN) & random(DGN) & ours(Objaverse) & random(Objaverse)\\ \midrule
Q1($\times10^{-2}$) &\textbf{ 1.25} & 0.47 &\textbf{1.41} & 0.79\\
\bottomrule
\end{tabular}
\caption{The region quality comparison between our method and random sample. The results demonstrate that our initialization strategy can initialize grasp region candidates with significantly higher quality.}
\label{tab:inference_time}
\end{table}

% \subsection{Qualitative Experiments of Decoupled Force Closure}
\begin{figure}[h]
    \centering
    \includegraphics[width=\linewidth]{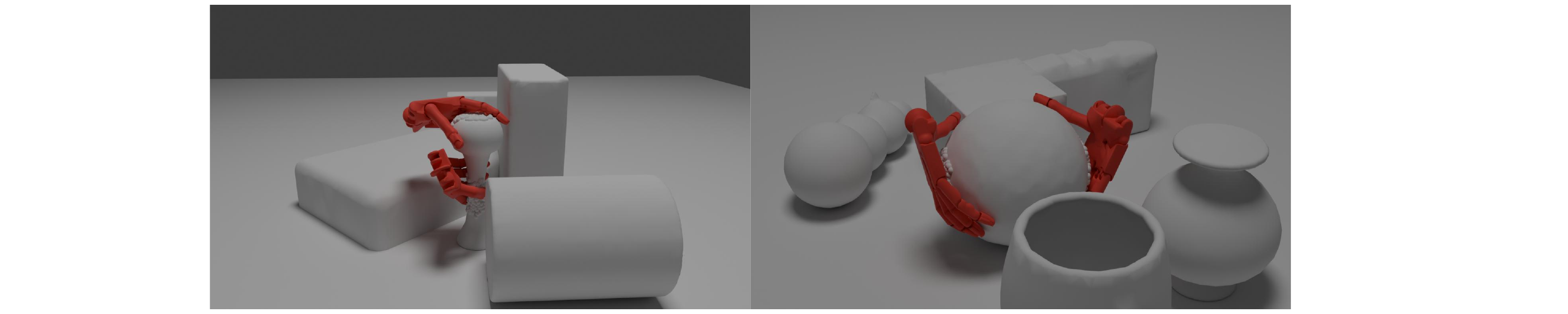}
    \caption{Visualization of extending our pipeline to cluttered scenes.}
    \label{fig:cluster}
\end{figure}

\subsection{Extension to Cluttered Scenes}

Our pipeline can be readily extended to cluttered environments (Fig.~\ref{fig:cluster}). By incorporating additional collision constraints, the method is able to optimize collision-free grasps in complex, cluttered scenarios.

\subsection{More Visualization of Diverse Grasping on Data Generation}

\subsubsection{Visualization of other Dexterous Hands}
\begin{figure}[h]
\centering
\includegraphics[width=\linewidth]{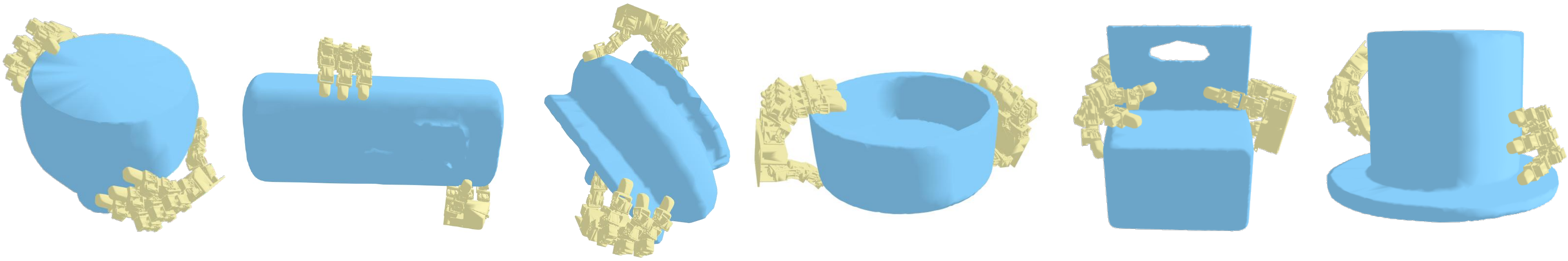}
% \vspace{-0.5cm}
\caption{The visualization of the reslut of our data synthesis on LeapHand. Our data synthesis pipeline can be used for different dexterous hands.
}
\label{fig: leaphand}
\end{figure}

Our data synthesis pipeline is not limited to a specific hand model and can be naturally extended to other dexterous hands. Fig.~\ref{fig: leaphand} visualizes grasp samples generated by applying our pipeline to the Leap Hand \cite{shaw2023leap}, demonstrating the generality of the proposed method across different dexterous hand embodiments.
\subsubsection{Visualization of Grasps in our Dataset}
\begin{figure}[h]
\centering
\includegraphics[width=\linewidth]{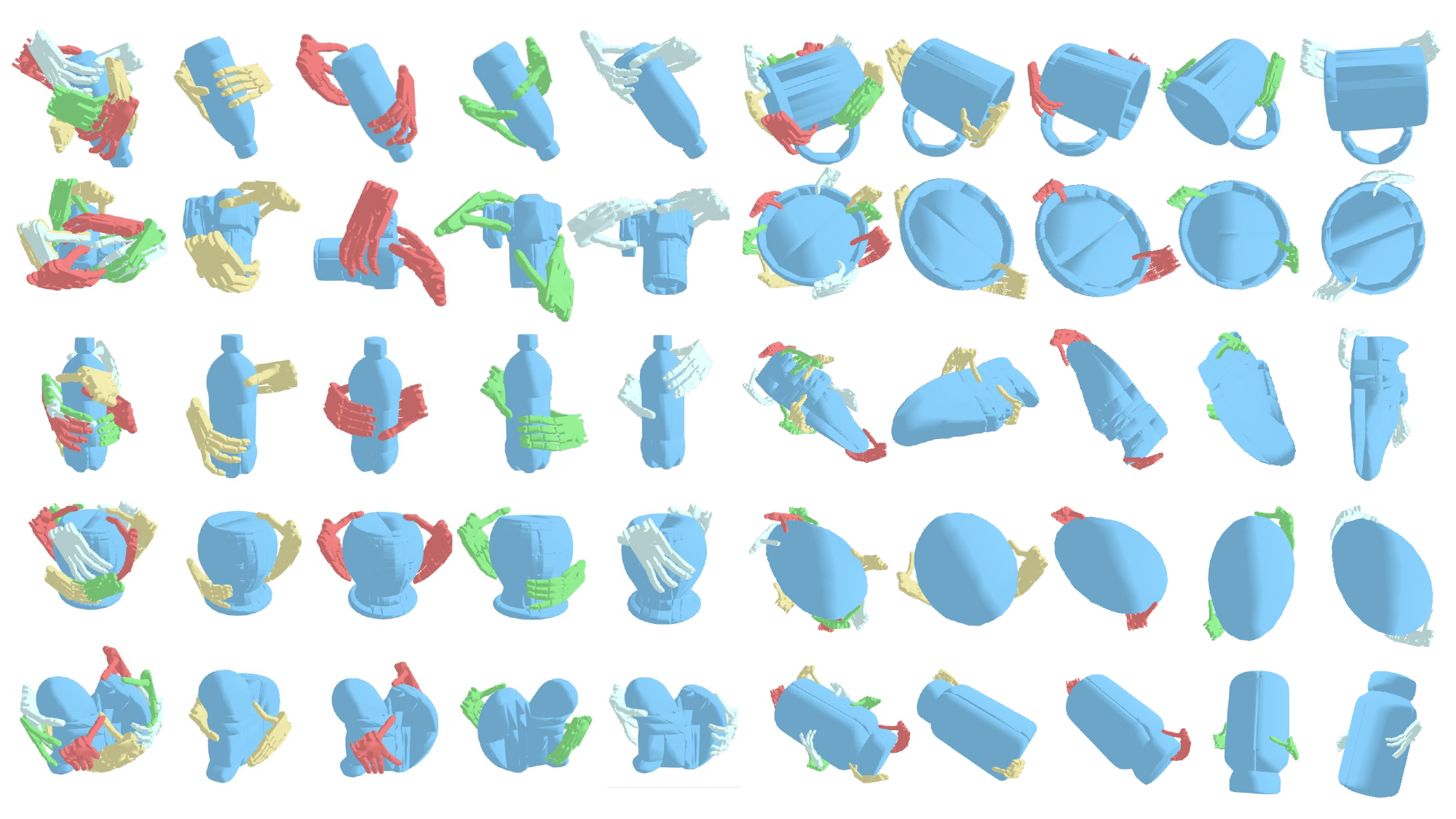}
% \vspace{-0.5cm}
\caption{Visualization of the dataset diversity. The samples span multiple object scales and demonstrate diverse grasping poses even under identical object placements.}
\label{fig:dataset_vis}
\end{figure}

We present a visualization of the grasps contained in our dataset in ~\ref{fig:dataset_vis}. These examples demonstrate that our dataset covers a wide range of object sizes and exhibits significant grasp diversity when objects are placed in identical poses on the tabletop.

\section{Experiments of Framework}
% \subsection{Details of Experiments}

\subsection{Reproduction of SOTA method}

\textbf{DGTR}\cite{xu2024dgtr} We utilize the original three independent MLP heads—responsible for predicting translation, rotation, and joint angles—by expanding their output dimensions to generate bimanual poses. The total loss is defined as the summation of the individual losses for both hands, with calculation metrics consistent with the original paper. Furthermore, we strictly adhere to the official three-stage training strategy.

\textbf{SceneDiffuser}\cite{huang2023scenediffuser} We doubled the input and output dimensions of the SceneDiffuser model to represent both left- and right-hand grasping, while continuing to use the previous loss function.

% \subsection{Experiments on other bimanual dexterous grasping datasets}

\subsection{Efficiency Analysis of Framework}
To further evaluate our model, we conduct ablation studies on inference time. We measure the inference time of a single batch with batch size 64 and compare our method against the DDPM-Baseline (removes the BCM, GAF and SAGA component). To reduce measurement variance, we run each experiment for five epochs and report the mean and standard deviation. As shown in Table~\ref{tab:inference_time}, and the results indicate that our method significantly improves the success rate at the cost of only a modest increase in inference time.
\begin{table}[htbp]
\centering
\begin{tabular}{lcc}
\toprule
Method & Average Inference Time (s) & Success Rate-Lift(\%)\\ \midrule
Baseline & $1.21 \pm 0.03$ & 41.17\\ 
% BimanGrasp-DDPM & 8.19 & --\\ \midrule
Ours & $1.67 \pm 0.05$ & 61.93\\
\bottomrule
\end{tabular}
\caption{Inference time comparison per batch. Our method achieves a good balance between model performance and efficiency.}
\label{tab:inference_time}
\end{table}

\subsection{Further Ablation of the Bimanual Coordination Module}

We further conduct an ablation study on the bimanual coordination module to analyze the view sampling strategy. In our full model, the coordinated grasp views are sampled in a conditional manner. Specifically, we first sample the primary grasping view from the predicted single-view probabilistic $p_i$. Given the sampled primary view, we then select the corresponding row from the predicted bi-view probabilistic $p^{bi}_{i,j}$ and sample the second view conditioned on the primary one.

For ablation, we remove this conditional sampling procedure and instead directly sample a single entry from the bi-view probabilistic $p^{bi}_{i,j}$. In this case, the sampled index jointly determines the primary and second grasping views without conditioning. Although this strategy treats the two hands in a fully symmetric manner, the results in Table~\ref{tab:bcm_abation} show a clear performance degradation across most metrics. This is because the conditional sampling in our design decomposes view selection into two sequential decisions, which is significantly easier than directly selecting a view pair from the full bi-view distribution, leading to more reliable bimanual coordination.

\begin{table}[htbp]
    \centering
    \begin{tabular}{l cccc} % 去掉 @{\extracolsep{\fill}}，改用标准 tabular
        \toprule
        \textbf{Method} & \multicolumn{4}{c}{\textbf{Metrics}} \\
        \cmidrule(l){2-5}
        & SUC-L$\uparrow$ & PD$\downarrow$ & SPD$\downarrow$ & CDC$\downarrow$ \\
        \midrule
        Direct Bi-view Sampling  & 45.49          & 2.23          & \textbf{0.01} & 2.96          \\
        Ours           & \textbf{61.93} & \textbf{1.73} & 0.02          & \textbf{2.56} \\
        \bottomrule
    \end{tabular}
    \caption{Further Ablation of the Bimanual Coordination Module. This module can significantly improve the framework performance.}
\label{tab:bcm_abation}
\end{table}

% \subsection{More visualization about Scale-Adaptive Grasp Anchor}
% 
% 
% \textbf{Reduction of the action space by Scale-Adaptive Grasp Anchor}

% \textbf{Visualization of the choosing of Scale-Adaptive Grasp Anchor}

\subsection{Failure Case Analysis}
\begin{figure}[h]
\centering
\includegraphics[width=\linewidth]{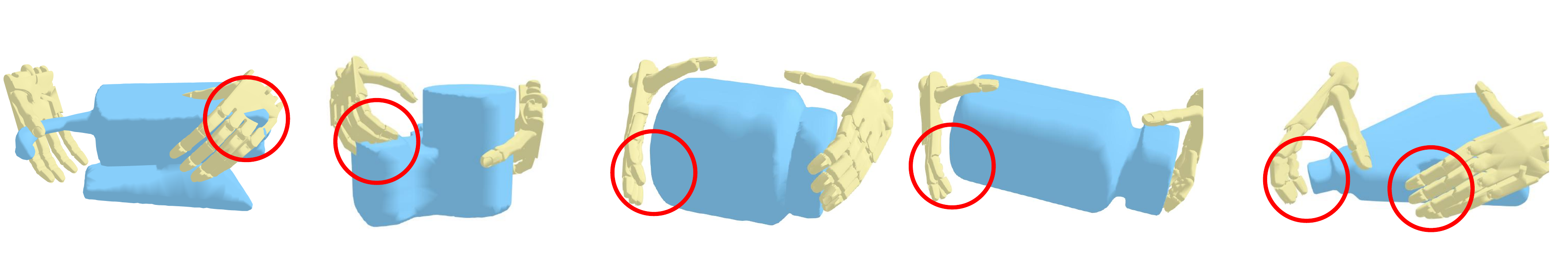}
% \vspace{-0.5cm}
\caption{The visualization of failure cases. We find that the common failure cases are caused by object-hand penetration and non-contact.
}
\label{fig: shadowfail}
\end{figure}

As illustrated in Fig.~\ref{fig: shadowfail} , failure cases primarily manifest as physical instability. Minor object penetration or floating grasps may occur when the generative model fails to perfectly balance the penetration penalty and contact supervision. Furthermore, as demonstrated in the rightmost case, a small number of objects undergo lateral sliding due to insufficient force closure, resulting in grasp failure.

\subsection{More Visulization of Generated Poses}
\begin{figure}[h]
\centering
\includegraphics[width=\linewidth]{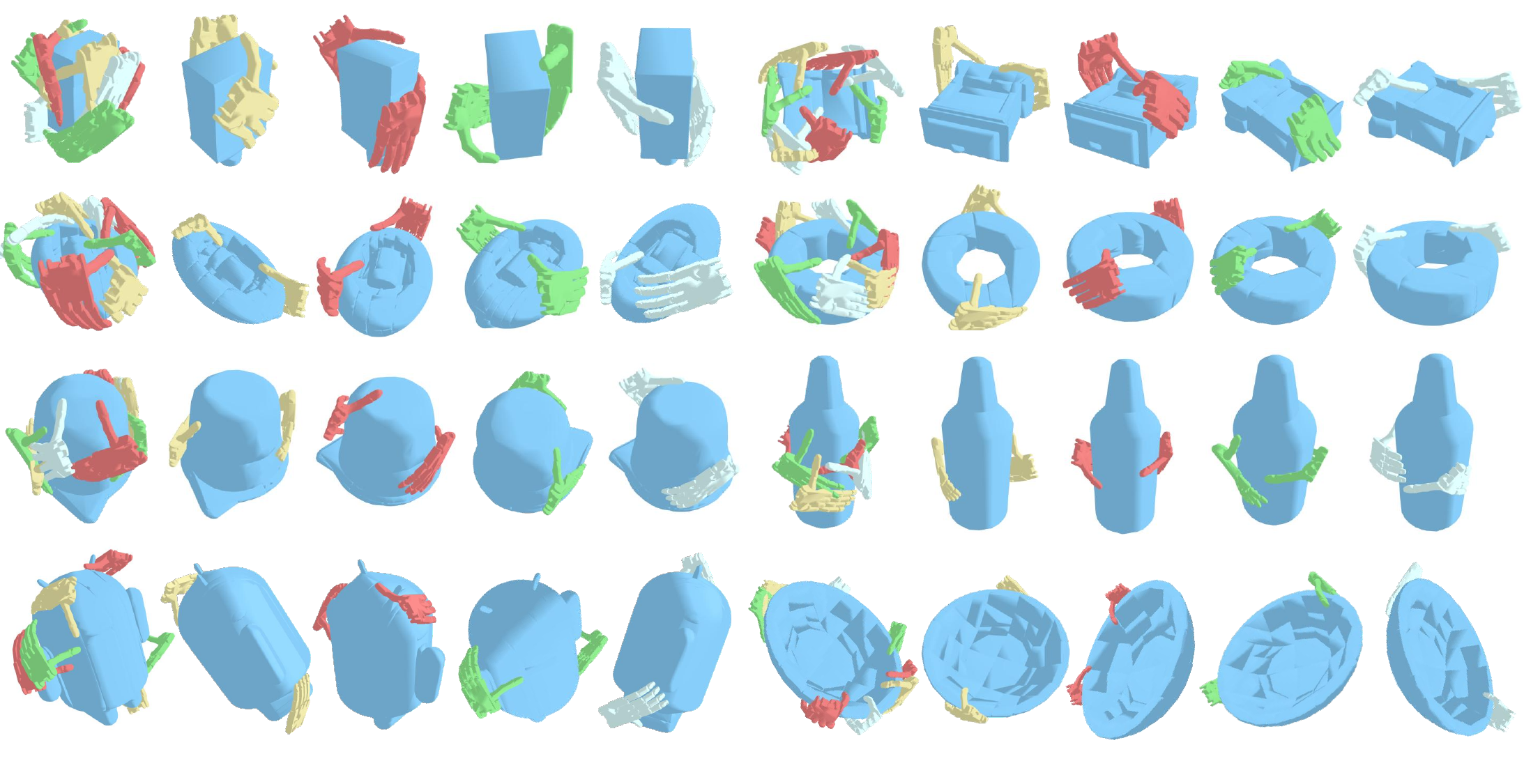}
% \vspace{-0.5cm}
\caption{The visualization of dexterous grasping generated by our model.
}
\label{fig: model_diversity}
\end{figure}

In Fig.~\ref{fig: model_diversity}, we provide additional visualizations of the generated bimanual grasps  on various objects. As observed, our framework consistently produces physically plausible and stable grasps, effectively adapting to objects with diverse geometric shapes and sizes. Furthermore, the model is capable of generating a rich variety of grasping postures for the same object, and the visualized results demonstrate distinct bimanual coordination.
% \clearpage
\section{Real world Experiments}
\subsection{Real world Experiments Details}
In this subsection, we describe the real-world experimental pipeline.For the single-view point cloud setting, we apply Grounded-SAM~\cite{ren2024groundedsam} to segment the object, and use the resulting mask to crop the object point cloud from the RGB-D frame as input to the model. For full points cloud setting, we first reconstruct the target object using a single-view 3D reconstruction method~\cite{hyper3d2024} and align the observed point cloud with the reconstructed mesh by matching their 3D bounding boxes to obtain the object scale. Next, we estimate the object 6D pose using 6D pose estimation~\cite{wen2024foundationpose} and we obtain the object point cloud input to our model by uniformly sampling points from the surface of the scaled mesh after world-frame pose transformation. 

We then convert the predicted Shadow Hand joint poses into the specific dexterous hand joint configuration via a hand-object consistent retargeting \cite{wei2025omnidexgrasp}. We define the pre-grasp pose by offsetting the final grasp pose along the hand’s negative approach direction (moving backward from the grasp along the hand-back/approach axis) by a fixed distance.

\begin{figure}[t]
\centering
\includegraphics[width=1.0\linewidth]{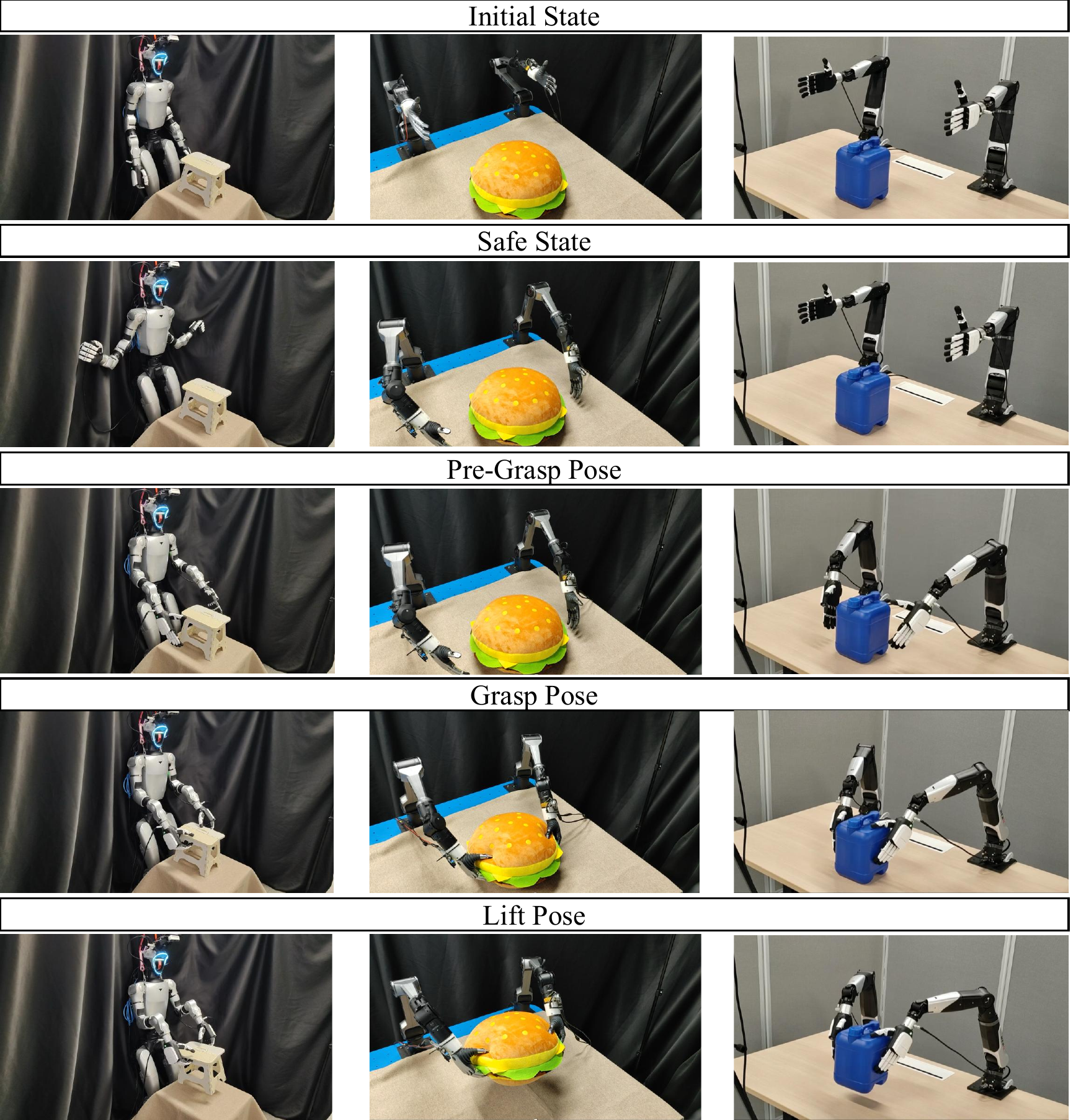}
% \vspace{-0.5cm}
\caption{The visualization of real world execution. During execution, the robot arm and hand first move from a predefined safe configuration to the pre-grasp pose. The system then commands the arm–hand to move from pre-grasp to the final grasp pose, and then to lift pose to complete the grasp.
}
\label{fig: real execution}
\end{figure}

During execution, as shown in Figure \ref{fig: real execution}, the robot arm and hand first move from a predefined safe configuration to the pre-grasp pose. The system then commands the arm–hand to move from pre-grasp to the final grasp pose, and then to lift pose to complete the grasp.

% \begin{figure}[h]
% \centering
% \includegraphics[width=\linewidth]{asset/realworl_pipeline.pdf}
% % \vspace{-0.5cm}
% \caption{The overall pipeline of real world deployment. Given a single-view RGB-D observation, we first reconstruct the target object into a mesh and estimate the object’s 6D pose in the camera frame. Then we employ our BiDexGrasp framework to generated grasp poses in object frame based on the sampling point cloud. This pose are transformed into the robot base frame using the object 6D pose and hand–eye transformation. And we employ a optimization-based retargeting to obtain the specific dexterous joints. And the corresponding pre-grasp pose is defined by offsetting the final grasp pose a fixed distance along the hand’s negative approach direction.
% }
% \label{fig: real deploy}
% \end{figure}

%%%%%%%%%%%%%%%%%%%%%%%%%%%%%%%%%%%%%%%%%%%%%%%%%%%%%%%%%%%%
\end{document}